\documentclass[iicol]{sn-jnl}

\usepackage[numbers]{natbib}
\usepackage{graphicx}%
\usepackage{multirow}%
\usepackage{amsmath,amssymb,amsfonts}%
\usepackage{amsthm}%
\usepackage{mathrsfs}%
\usepackage[title]{appendix}%
\usepackage{xcolor}%
\usepackage{textcomp}%
\usepackage{manyfoot}%
\usepackage{booktabs}%
\usepackage{algorithm}%
\usepackage{algorithmic}
\usepackage{listings}%
\usepackage{array}
\usepackage{stfloats}
\usepackage{verbatim}
\usepackage{enumitem}
\usepackage{bm}
\usepackage{float}
\begin{document}
\title[Article Title]{CEI-3D: Collaborative Explicit-Implicit 3D Reconstruction for Realistic and Fine-Grained Object Editing}


\author{Yue Shi}\email{shiyue001@sjtu.edu.cn}

\author{Rui Shi}\email{shi-rui@sjtu.edu.cn}

\author{Yuxuan Xiong}\email{xiongyx@sjtu.edu.cn}

\author*{Bingbing Ni\textsuperscript{*}} \email{nibingbing@sjtu.edu.cn}
\author{Wenjun Zhang}\email{zhangwenjun@sjtu.edu.cn}


\affil{\orgdiv{School of Electronics, Information and Electrical Engineering}, \orgname{Shanghai Jiao Tong University}, \orgaddress{\street{Dongchuan Road 800}, \city{Shanghai}, \postcode{200240},\country{China}}}


\abstract{3D reconstruction and editing from 2D images have important applications in creative content generation, customization, and interactive design. However, existing 3D editing methods often produce unrealistic and unrefined results due to the deeply integrated nature of their reconstruction networks. To address the challenge, this paper introduces CEI-3D, an editing-oriented reconstruction pipeline designed to facilitate realistic and fine-grained editing.
Specifically, we propose a collaborative explicit-implicit reconstruction approach, which represents the target object using an implicit SDF network and a differentially sampled, locally controllable set of handler points. The implicit network provides a smooth and continuous geometry prior, while the explicit handler points offer localized control, enabling mutual guidance between the global 3D structure and user-specified local editing regions. To independently control each attribute of the handler points, we design a physical properties disentangling module to decouple the color of the handler points into separate physical properties. We also propose a dual-diffuse-albedo network in this module to process the edited and non-edited regions through separate branches, thereby preventing undesired interference from editing operations. Building on the reconstructed collaborative explicit-implicit representation with disentangled properties, we introduce a spatial-aware editing module that enables part-wise adjustment of relevant handler points. This module employs a cross-view propagation-based 3D segmentation strategy, which helps users to edit the specified physical attributes of a target part efficiently. 
Extensive experiments on both real and synthetic datasets demonstrate that our approach achieves more realistic and fine-grained editing results than the state-of-the-art (SOTA) methods while requiring less editing time. Our code is available on \href{https://github.com/shiyue001/CEI-3D}{https://github.com/shiyue001/CEI-3D}.}


\keywords{3D Reconstruction, 3D Editing, Collaborative Explicit-Implicit Representation}



\maketitle

\section{Introduction}\label{sec1}

\begin{figure}[t]
\centering
\includegraphics[width=.95\linewidth]{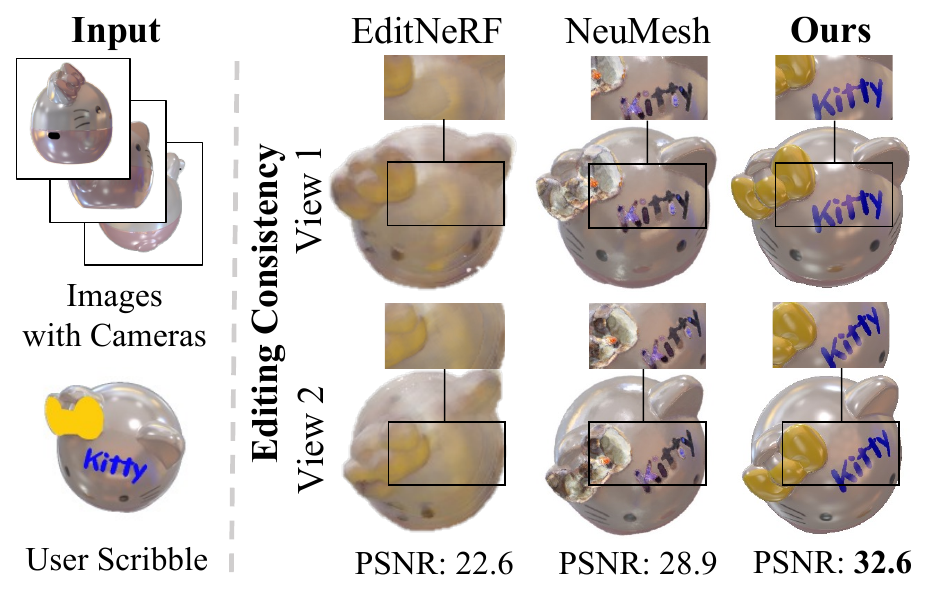}
\caption{\small Comparison with popular methods. Our approach shows better editing results. It maintains the photo-realism and fidelity to the user scribble when synthesized from different viewpoints. Not only is our PSNR (dB) improvement encouraging, but the reduction in edit time is impressive. Concretely, editing times on this example of EditNeRF \citep{liu2021editing}, NeuMesh \citep{yang2022neumesh}, and Ours are 45s, 1h, and 14s.}
\label{fig:intro-img}
\end{figure}

Given multi-view 2D images and the single-view user scribble, 3D editing methods \citep{liu2021editing, yang2022neumesh, NeRF-editing, kuang2023palettenerf} reconstruct the target object, modify it and display the edited object by rendering from novel viewpoints. Recently, implicit 3D representations \citep{nerf, park2019deepsdf} demonstrate remarkable capabilities in 3D reconstruction and novel view synthesis, producing photo-realistic rendering results. 
However, implicit 3D representations inherently lack the flexibility for editing, such as modifying object appearance or geometry, which limits their applicability in 3D interaction, generation, and artistic creation for general users. While some approaches enable editing by introducing a controlling branch \citep{liu2021editing} or integrating mesh representation \citep{yang2022neumesh}, these methods often fall short in terms of accuracy, refinement, and naturalness of the results (refer to Fig.~\ref{fig:intro-img}). The primary challenges in editing implicit 3D models stem from three aspects.

\textbf{Challenge 1.} \textit{The entire object is reconstructed by a continuous function (neural network), which inherently does not support localized modifications.} Implicit representations \citep{park2019deepsdf, nerf} encode the target object using a neural network and enable continuous queries, showing advantages in realistic rendering \citep{nerf, park2019deepsdf, mescheder2019occupancy}, scalability \citep{tancik2022block, xie2021fig}, scene agnostic behavior \citep{graf, wangibrnet}, and etc. However, implicit representations lack flexibility. Specifically, when a local modification is made, the change in global information will inevitably cause changes in the non-edited area (refer to Section \ref{ssec: preliminaries}  for further analysis). To improve the editability of implicit representations, EditNeRF \citep{liu2021editing} implements object editing by introducing a modulation network component and uses a large amount of category data to implement part editing. Due to the dependence on the learned prior information, EditNeRF has a poor generalization to arbitrary objects and does not support precise local pattern painting. For instance, when EditNeRF is used to scribble in Fig.~\ref{fig:intro-img}, the scribble color spills over the whole object. More recently, NeuMesh \citep{yang2022neumesh} combines mesh and neural network representation methods to improve the editing flexibility of implicit representations.
Specifically, NeuMesh encodes the neural implicit field with geometry and
texture codes on mesh vertices. Then, it realizes editing by updating the neural implicit field and the corresponding codes.
Although NeuMesh surpasses EditNeRF on editing capability, it exhibits coarse edge patterns and inaccurate colors that stem from the limited resolution of its mesh representation. Besides, it uses the Marching Cube algorithm \citep{lorensen1998marching} to extract a mesh from the implicit reconstruction network. This process increases time consumption and introduces errors due to the geometric discretization involved in transforming continuous representations into the discrete mesh form. NeuMesh takes over 30 hours for training and over 1.7 hours for editing, which is computationally expensive. The limitations of current methods suggest that collaboratively integrating high-fidelity, easily reconstructed continuous surfaces with localized, fine-grained control points may offer a promising direction to achieve both high rendering quality and flexible editing capability.

\textbf{Challenge 2.}
\textit{Effective realistic editing often requires alterations in specific attributes, yet most implicit editing models can not separately model lighting and material properties.} The color of the object in real scenes is physically formed by the combination of lighting and the attributes (including geometry and material) of the object. However, the mentioned editing approaches above use RGB to represent the object color. When editing the object color, they will change the lighting and the object attributes, which affects the realism of the object editing. As shown in the second column of Fig.~\ref{fig:intro-img}, the neglect of color composition often results in inferior results for personalized scribbles. The editing appears superficially overlaid on the image, lacking realism and making it look like they are not part of the original object. Recent studies \citep{li2025recap,wu2025deferredgs, zhang2021nerfactor, physg2021} explore lighting estimation and scene relighting. However, these approaches cannot be directly applied to editing because of the limited granularity and controllability of the representation.  Therefore, there is an urgent need for an editing method that reconstructs the target object by disentangling lighting and attributes at a fine-grained level to facilitate natural editing.


\textbf{Challenge 3.} \textit{Existing editing methods have weak spatial awareness and they require multi-view editing to change the specified parts of an object.} In fact, a single viewpoint scribble can describe the parts that the user expects to edit, which enables more efficient editing. The most important reason for the weak spatial perception of existing editing methods \citep{yang2022neumesh} is that they cannot obtain 3D segmentation labels. These labels segment parts from an object and thus establish the correspondence between multiple views of the parts. For example, when users want to edit the entire ear of a kitty model, scribbling from a single view only affects a portion of the ear. To complete the task, users must scribble from multiple viewpoints. However, coloring one part of the 3D ear at a time is neither efficient nor convenient. 
Recently, some works investigate spatial-aware editing. For instance, EditNeRF proposes conditional radiation fields to learn part information from 3D category datasets. It must be trained on class-specific datasets, which are scarce, so it cannot generalize to arbitrary objects. PaletteNeRF \citep{kuang2023palettenerf} segments the target object by learning the 3D feature field from the image-based segmentation model Lang-Seg \citep{li2022language}. However, since it represents the scene by a weighted summation of basic colors, PaletteNeRF cannot decompose multiple parts with similar colors, which causes unwanted color changes in non-edited areas.
Besides, these models do not support interactive modifications when segmentation is unsatisfactory for users.
Thus, existing editing methods are not suitable for real-time and reasonable 3D parts editing.

\begin{figure}[t]
\centering
\includegraphics[width=.95\linewidth]{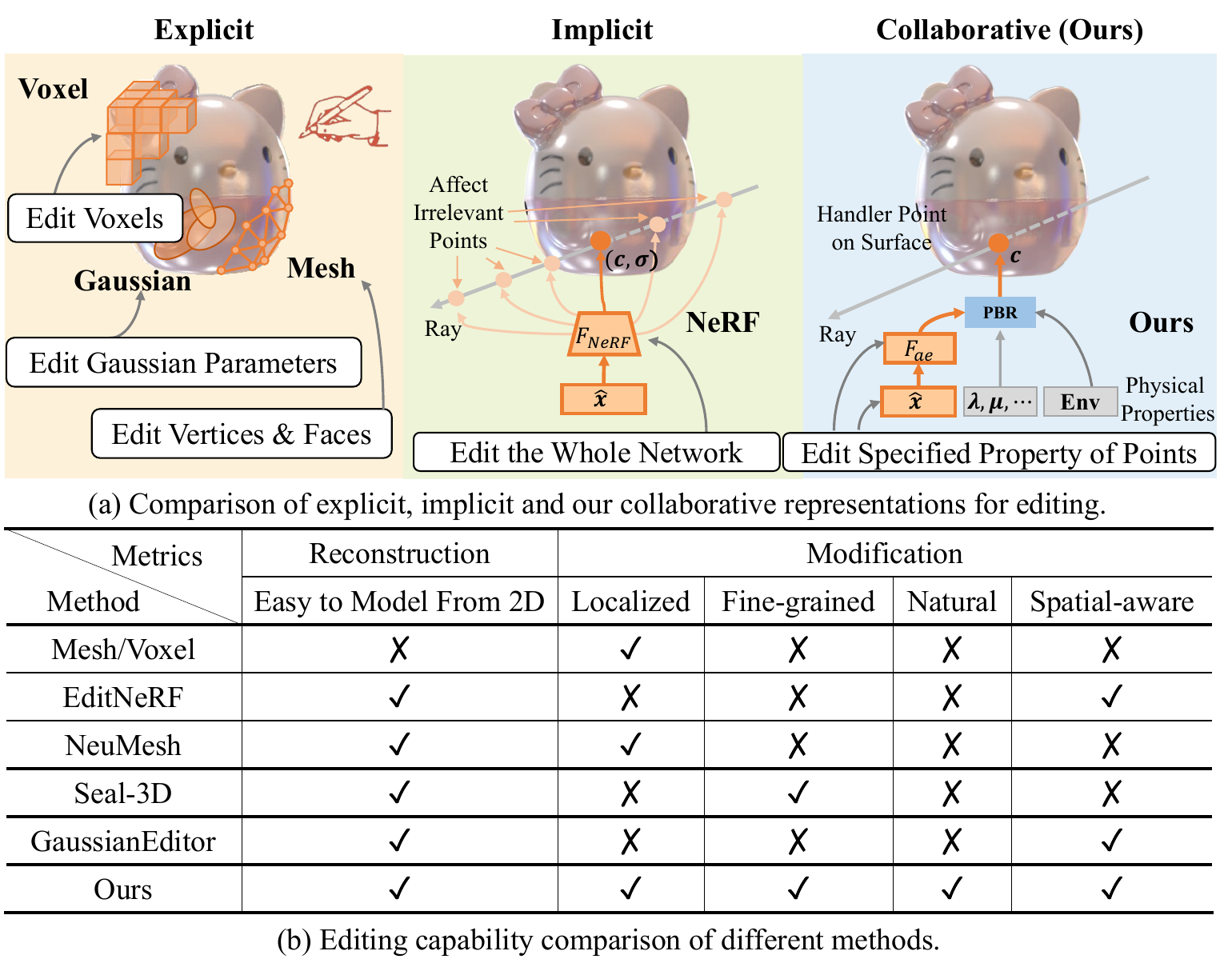}
\caption{\small The main ideas and the capabilities comparison. (a) Unlike traditional explicit or implicit methods, our approach facilitates editing by introducing flexible handler points and disentangling physical properties. (b) In comparison, our approach satisfies properties that are desirable for editing.}
\label{fig:intro-img2}
\end{figure}

To address these challenges, we propose an editing-oriented 3D reconstruction framework, \textbf{C}ollaborative \textbf{E}xplicit-\textbf{I}mplicit \textbf{3D} Reconstruction framework (\textbf{CEI-3D}), which enables realistic and fine-grained object editing by reconstructing the 3D object in an inherently editable form. We compare the main ideas and capabilities of existing methods with ours in Fig. \ref{fig:intro-img2}. CEI-3D reconstructs the object into a collaborative explicit-implicit representation. By adjusting the attributes of localized explicit handler points, users can perform fine-grained and controllable edits that drive changes in the implicit surface. This design retains the high-fidelity rendering and faithful reconstruction advantages offered by implicit representations, while introducing the editing flexibility and fine-level controllability that are typically afforded by explicit structures. Building on this controllability, CEI-3D further enhances realism by disentangling physical properties such as lighting and material, enabling fine-grained control over both spatial locations and physical properties. Specifically, CEI-3D consists of three modules: \textbf{EIR} (\textbf{E}xplicit-\textbf{I}mplicit \textbf{R}econstruction), \textbf{PPD} (\textbf{P}hysical \textbf{P}roperties
\textbf{D}isentangling module), and \textbf{SAE} (\textbf{S}pace-\textbf{A}ware \textbf{E}ditor).
We establish the EIR module to reconstruct 3D objects to obtain a collaborative representation that contains implicit networks and explicit handler points. The handler points, which are sampled from the continuous implicit surface using a differentiable sampling strategy, are editable and address the localized modification challenges.
Based on the reconstructed 3D representation, we further propose PPD to disentangle environment lighting and object attribute parameters of handler points, preparing for realistic editing.
PPD defines learnable disentangled features and network branches to represent the disentangling properties, which are combined to form the rendered color through the physically based rendering equation.
By disentangling, the specified attribute network can be modified to make the scribble as realistic as it was born. 
Lastly, we propose SAE to assign a part label for each handler point, which realizes spatial-aware editing. Part labels are obtained by our proposed Cross-View Propagation-Based 3D Segmentation (CPS) strategy, which generates 2D masks using a 2D pre-trained Segmentation Anything Model (SAM) \cite{sam} and back-projects them onto the 3D objects. Considering the misclassified problem of part labels, we design a prompt based on positive and negative samples to guide SAM to perform more detailed segmentation. With the help of part labels, SAE edits the specified part of an object using the single-viewpoint scribble.

Extensive experimental results demonstrate that our approach achieves realistic editing without color breaks or bleed like others (see Fig.~\ref{fig:intro-img}). 
The primary contributions are as follows:
    \begin{itemize}[]
    \item We propose a collaborative explicit-implicit reconstruction framework that represents 3D objects in an inherently editable form. Fine-grained and controllable editing is enabled by manipulating explicit handler points, while global surface fidelity is preserved through the implicit representation. The explicit handler points influence the implicit surface through a dual-branch design, where only regions marked as editable are updated.
    \item Building on the reconstructed collaborative representation, we design a physical properties disentangling module to facilitate the targeted modification of specified attributes, enabling fine-grained control over both spatial regions and physical properties, and thereby improving editing realism.
    \item Leveraging the above training results, we propose a spatial-aware editing module that enables region-level editing from a single-viewpoint scribble. Part labels are assigned to handler points via our proposed cross-view propagation strategy, which utilizes 2D masks derived from SAM, with a prompt mechanism to align the segmentation with the user intended part decomposition.
\end{itemize}

The rest of this paper is organized as follows. In Section \ref{relatedwork}, we discuss the related works. Section \ref{ssec: preliminaries} introduces the preliminary and problem formulation. Section \ref{method} provides a detailed description of our method. Section \ref{experiments} shows the results of empirical studies and ablation experiments. Concluding remarks are provided in Section \ref{conclusion}.

\section{Related Work}
\label{relatedwork}
Existing 3D editing methods consist of two main techniques, i.e., 3D reconstruction from 2D multi-view images and 3D editing based on user scribbles. Therefore, we introduce the related methods from the following two aspects:

\noindent
\textbf{3D Reconstruction.}
3D reconstruction is the process of inferring 3D geometric shapes from 2D multi-view images, which requires accurately capturing the structure and appearance of the target object from different perspectives. Traditional 3D reconstruction methods usually rely on explicit 3D shape representations. Mesh is the most widely used explicit 3D representation, dominating various 3D content creation applications \citep{xu2024instantmesh,zhao2025hunyuan3d,xiang2025structured}. Numerous sophisticated software tools have been developed to support mesh manipulation, including Blender, Houdini, and ZBrush \citep{zbrush2023}. Although powerful, modeling of 3D mesh requires considerable technical skills and detailed manual effort for accurate modeling and editing, which is challenging for users without specialized training. Besides, mesh represents the object by the discrete vertices and edges \citep{Shi_2021_ICCV, p2m++}, which are difficult to optimize with neural networks directly from images \citep{wang2021neus}.

Recently, as the demand for high-quality rendering and scalable modeling increases, implicit 3D  representations\footnote{Readers interested in different types of 3D data representations and their benefits may refer to the recent survey paper by Samavati et al. \citep{samavati2023survey}.} \citep{ park2019deepsdf, nerf, muller2022instant, kerbl20233dgs, mello2025neural} become prevalent in 3D reconstruction. By modeling geometry and appearance as continuous functions, implicit methods \citep{caselles2025implicit, SHI2025102996, lin2025patch} capture fine details and optimize directly from images, enabling photorealistic rendering and seamless integration with neural networks. In particular, NeRF \citep{nerf} represents scenes as volumetric opacity fields via multilayer perceptrons (MLPs) by fitting input views and images. Surface-based neural rendering methods represent appearance by an MLP that takes 3D points and viewing directions and outputs a color \citep{wang2023neus2, schirmer2024surveyofsdf}. These methods work well for novel-view rendering but are unsuitable for editing due to the network's tightly coupled nature, making local modifications challenging. In contrast to these methods, we model the object in a spatial and attribute-disentangling manner, facilitating fine-grained and realistic editing.

\noindent
\textbf{3D Editing.} Based on the reconstructed 3D object, 3D editing is employed for various applications, including scribbling, geometric deformation, and other modifications, enabling users to customize their 3D models effectively. To realize 3D editing, some text-guided methods \citep{chen2024gaussianeditor, li2024focaldreamer, zhuang2023dreameditor} modify the 3D scene or object with the help of a stable diffusion model \citep{rombach2021stablediff}. These methods enable creative, coarse-grained edits but lack precise and user-friendly controls for scribble-based modifications. Some methods \citep{CCNERF, 2023neuraleditor, NeRF-editing} are limited to rigid transformations or object-level modifications, which are not suitable for fine-grained local pattern drawing. Some lighting estimation methods \citep{zhu2023i2sdf, zhang2023neilf++23, li2025recap} are capable of simulating and changing global light or material. Yet, they only achieve coarse editing granularity and do not support fine-grained editing. 
More recently, ICE-NeRF \citep{lee2023ice}, Sine \citep{bao2023sine} and
Palette-NeRF \citep{kuang2023palettenerf} supports part-aware color changing using knowledge from the 2D segmentation models. However, they produce results at a coarse granularity, treating object as a whole without identifying specific parts. Besides, none of these methods support fine-grained local pattern editing.

The most related works are EditNeRF \citep{liu2021editing}, NeuMesh \citep{yang2022neumesh}, Seal3D \citep{wang2023seal}, and GaussianEditor \citep{chen2024gaussianeditor}. EditNeRF introduces a modulation network to facilitate object editing by leveraging category data for part editing. However, it relies on the learned prior information, which limits its generalization to arbitrary objects, leading to color spilling over the entire object. NeuMesh encodes the color information to mesh vertices and realizes editing by modifying the corresponding codes. While NeuMesh achieves pattern painting, it makes color mistakes in pattern edges. GaussianEditor \citep{chen2024gaussianeditor} achieves editing by fine-tuning the Gaussian parameters based on edited images. Although GaussianEditor produces reasonable results when supervised with multi-view edited images, it suffers from bright spots and color overflow on fine-grained user scribbles. This is primarily due to the non-uniformity of the Gaussian distribution and the imbalance in optimization when only a single view is available. Seal3D \cite{wang2023seal} introduces a proxy function for pattern mapping, but it struggles to accurately identify the edited surface, leading to imprecise modifications. This issue arises from its reliance on a volumetric representation, which hampers the precision of the editing. Besides, the existing approaches lack realism because they overlook the physics of rendering and directly alter the RGB values. These limitations highlight the difficulty of achieving both rendering quality and editing controllability within a representation. This motivates our exploration of a collaborative explicit-implicit representation that leverages the strengths of both paradigms—explicit structures for precise, user-controllable editing, and implicit fields for high-fidelity reconstruction and rendering.

\section{Preliminaries}
\label{ssec: preliminaries}
This section presents the problem formulation details and explains how to reconstruct and edit the object. 
Given a set of 2D multi-view images $ \mathcal{I} = \{ I_1, I_2, \ldots, I_N \}$ with camera parameters $\mathcal{C} = \{ C_1, C_2, \ldots, C_N \}$, where $I_i$ is the $i$-th image and $C_i$ is the corresponding camera parameter. Our target is to reconstruct the 3D object in an editing-oriented manner from multi-view images and edit it according to user scribbles. The related preliminaries are introduced as follows.

\noindent
\textbf{SDF-based 3D Reconstruction.} SDF is a widely used implicit 3D representation, which has the advantage of capturing intricate details to achieve highly realistic rendering.
SDF is a continuous function, generally modeled by a geometry network $f(\boldsymbol{x}; \Phi)$ and a radiance network $c(\boldsymbol{x}, \boldsymbol{d}; \Upsilon)$, where $\boldsymbol{x}\in \mathbb{R}^{3}$ represents the coordinate of a 3D point $p$. The function $f(\boldsymbol{x}; \Phi)$ outputs the signed distance from the point $p$ to the nearest surface of the object. $c(\boldsymbol{x}, \boldsymbol{d}; \Upsilon)$ predicts the view-dependent color $\boldsymbol{c}\in\mathbb{R}^{3}$ conditioned on coordinate $\boldsymbol{x}$ and
viewing direction $\boldsymbol{d}\in\mathbb{R}^{3}$. The $\Phi$ and $\Upsilon$ are corresponding network parameters. Different from NeRF \cite{nerf}, which lacks a clear definition of the surface, SDF clearly defines the object surface as the zero-level set $\mathcal{S}_{\Phi}$ of $f(\boldsymbol{x}; \Phi)$ as follows:
\begin{align}
\label{eq:sdf}
  \mathcal{S}_{\Phi}=\left\{\boldsymbol{x} \mid f(\boldsymbol{x} ; \Phi)=0\right\}.
\end{align}

SDF is rendered to images by volume rendering as proposed in \cite{wang2021neus}. Specifically, a ray \( \boldsymbol{r}(t) = \boldsymbol{o} + t \boldsymbol{d} \) is cast from the camera origin \( \boldsymbol{o}\in \mathbb{R}^{3} \) along the direction \( \boldsymbol{d}\) through each pixel. $t\in\mathbb{R}$ is the distance along the ray direction. Points along this ray are sampled, and their distances to the surface are evaluated using \( f(\boldsymbol{x}; \Phi) \). The color $\hat{\boldsymbol{c}}(\boldsymbol{r})\in\mathbb{R}^{3}$ for each pixel is obtained by accumulating color and opacity along the ray as $\hat{\boldsymbol{c}}(\boldsymbol{r}) = \sum_{i=1}^{P} T_i \varrho_i \boldsymbol{c}_i$,
where the transmittance $T_i = \exp \left( -\sum_{j=1}^{i-1} \varrho_j \delta_j \right) $ denotes the transmittance of sampled point. $\delta_j$ is the distance between neighboring sampled points. $P$ is the number of sampled points. $\varrho_i$ is the opacity calculated by the signed distance as in \cite{wang2021neus}.
SDF networks are optimized using the distance loss between the ground-truth pixel color $\boldsymbol{c}(\boldsymbol{r})$ and the predicted pixel color $\hat{\boldsymbol{c}}(\boldsymbol{r})$. 

In our method, we propose a collaborative explicit-implicit representation to make traditional SDF-based 3D reconstruction pipeline editable, enabling fine-grained and controllable editing over continuous implicit surfaces. We construct the geometry using an SDF network \cite{wang2023neus2} and then sample points on the implicit surface, as detailed in Section \ref{ssec: EIR}. Instead of directly using RGB colors, we optimize physical properties as in Section \ref{ssec:SADM}, to achieve realistic editing based on the proposed collaborative representation.

\noindent
\textbf{PBR and Disney BRDF.} Though RGB representation is widely used in SDF, the color we observe in the real world is the interaction result of geometry, lighting, and material properties. Physically Based Rendering (PBR) is designed to simulate this process. 
For the surface point $p$ with coordinate $\boldsymbol{x}$ and normal $\boldsymbol{n}$, we define that $L_{i}(\boldsymbol{\omega}_{i}; \boldsymbol{x})$ is the incident light intensity at location $\boldsymbol{x}$ along the direction $\boldsymbol{\omega}_{i}$, and $f_{r}\left( \boldsymbol{\omega}_{i}, \boldsymbol{\omega}_{o};\boldsymbol{x}\right)$ is the reflectance coefficient, where $\boldsymbol{\omega}_{i}$ and $\boldsymbol{\omega}_{o}$ denote the directions of incident lighting and reflected lighting. 
The color at coordinate $\boldsymbol{x}$ observed along the ray $\boldsymbol{r}$ with direction $\boldsymbol{d}$ ($\boldsymbol{d}=-\boldsymbol{\omega}_{o}$) is computed as follows:
\begin{equation}
\!\!\!\boldsymbol{c}(\boldsymbol{r})=L_{e}+\!\!\int_{\boldsymbol{\Omega}}\!\!L_{i}\left(\boldsymbol{\omega}_{i}\right) f_{r}\left( \boldsymbol{\omega}_{i}, \boldsymbol{\omega}_{o}; \boldsymbol{x}\right)\left(\boldsymbol{\omega}_{i} {\cdot} \boldsymbol{n}\right) \mathrm{d} \boldsymbol{\omega}_{i},\!\!
\label{eq: pbr}
\end{equation}
where $L_{e}$ represents the light emitted by the surface point $\hat{\boldsymbol{x}}$ itself, which satisfies $L_e = 0$ for non-luminous objects. The second term of Eq. (\ref{eq: pbr}) is an integral over all directions on a hemisphere $\Omega$ centered at $\boldsymbol{n}$, which calculates the total reflected light at the surface from various incident directions.
We use the Disney BRDF \cite{disneybrdf} to model the reflectance coefficient $f_{r}\left( \boldsymbol{\omega}_{i}, \boldsymbol{\omega}_{o}; \boldsymbol{x}\right)$ as the sum of the diffuse term $f_{d}$ and the specular term $f_{s}$ as follows:
\begin{align}
f_{r}\left( \boldsymbol{\omega}_{i}, \boldsymbol{\omega}_{o}; \boldsymbol{x}\right)=f_{d}\left( \boldsymbol{x}\right)+f_{s}\left( \boldsymbol{\omega}_{i}, \boldsymbol{\omega}_{o}; \boldsymbol{x}\right),
\label{eq:add f}
\end{align}
\begin{align}
f_{d}\left(\boldsymbol{x}\right)=\left (1-m\right) \frac{\boldsymbol{a}}{\pi},
\label{eq: fd}
\end{align}
\begin{align}
f_{s}\left( \boldsymbol{\omega}_{i}, \boldsymbol{\omega}_{o}; \boldsymbol{x}\right)=\frac{DFG}{4\left (\boldsymbol{\omega}_{i} {\cdot} \boldsymbol{n}\right)
\left(\boldsymbol{\omega}_{o} {\cdot} \boldsymbol{n}\right)}.
\label{eq: fs}
\end{align}
where $m\in[0,1]$ represents the metalness and $\boldsymbol{a}\in[0,1]^{3}$ is the diffuse albedo. $D$ is responsible
for the shape of the specular peak, $F$ is the Fresnel reflection coefficient, and
$G$ represents the shadowing factor. 
To make the editing realistic, we further simplify the PBR modeling of Eq. (\ref{eq: pbr}) into a closed form and derive disentangled physical properties in Section \ref{ssec:SADM}.  

\noindent
\textbf{3D Editing and Problems.} 3D editing is to modify the specified region of the reconstructed object according to the given user scribbled image $I_{e}$. The direct method for editing is to update the parameters of network $c(\boldsymbol{x}, \boldsymbol{d}; \Upsilon)$, which inevitably leads to undesired changes in non-edited regions. The current SDF is not suitable for editing due to its coupling in both geometry and attribute space. On the one hand, the network $f(\boldsymbol{x}; \Phi)$ and $c(\boldsymbol{x}, \boldsymbol{d}; \Upsilon)$ do not allow locally modifications. On the other hand, direct operations on the RGB color lead to unrealistic results. Besides, single-view scribbles cannot efficiently modify the whole related part. Therefore, we propose an editing-oriented framework by introducing handler points in Section \ref{ssec: EIR}, disentangling attributes in Section \ref{ssec:SADM} and spatial-aware editing in Section \ref{ssec:SEM}.

\section{Methodology}
\label{method}

\begin{figure*}[t]
\centering
\includegraphics[width=1\linewidth]{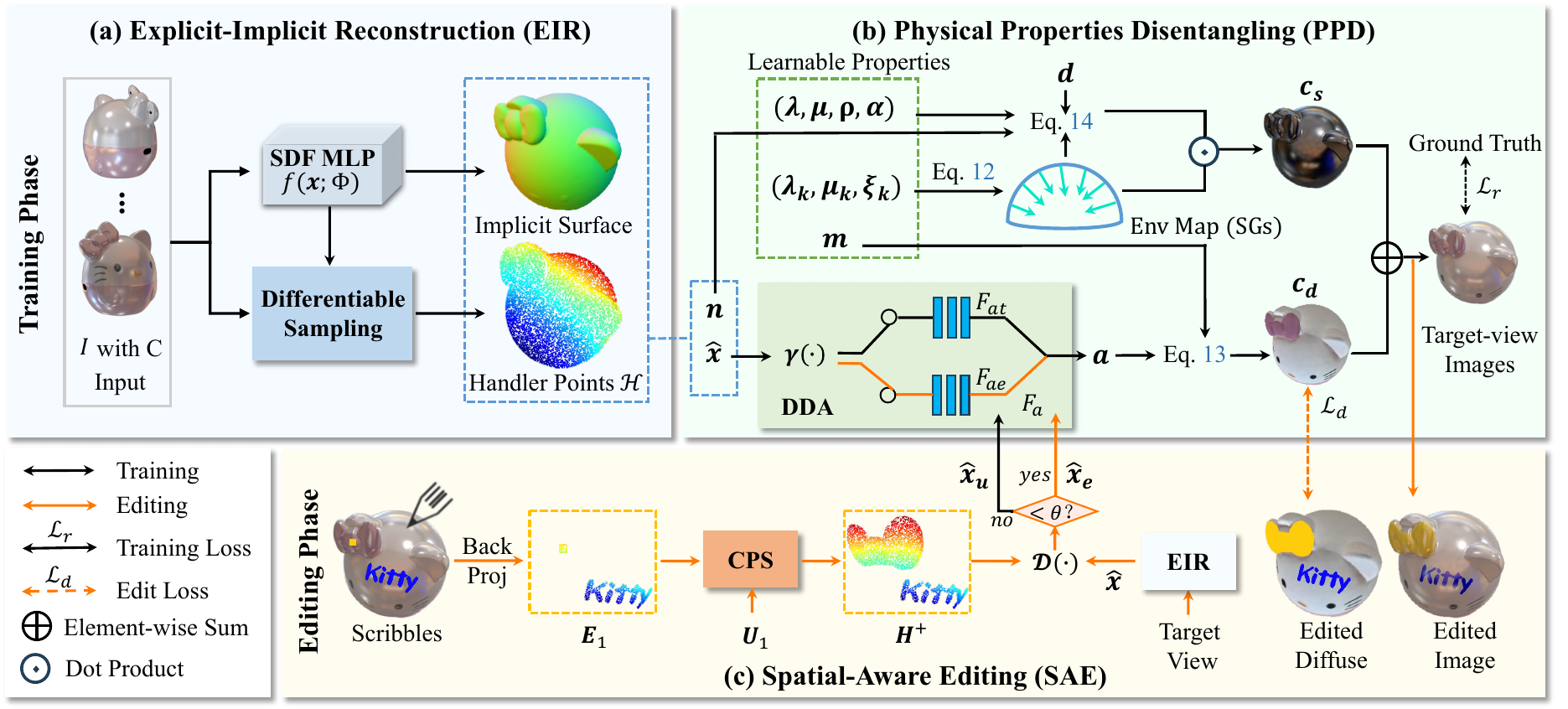}
\caption{\small Pipeline of CEI-3D. \textbf{Training Phase.} Given images and camera parameters, we first train an implicit surface network using an SDF MLP, shown on the top left of the diagram. To allow for flexible user control, we construct a handler point set $\mathcal{H}$ by sampling points on the implicit surface in a differential manner. Then, the combination of implicit surface and explicit handler points constitutes the collaborative 3D representation. To achieve controllability at the attribute level, we disentangle physical properties by optimizing learnable material parameters, lighting parameters, and diffuse branch $F_{at}$ for non-edited regions in the DDA network. \textbf{Editing Phase.} Given the user scribble, we find points to be modified and update the edited branch $F_{ae}$ of the DDA network using the target attribute values of specified points. After network parameter optimization, the edited results of any view can be rendered.  
}
\label{framework}
\end{figure*}
In this paper, we propose a collaborative explicit-implicit reconstruction framework (CEI-3D) to address the problem of inaccurate and unnatural 3D editing. By combining the high-fidelity reconstruction ability of implicit fields with the localized controllability of explicit handler points, CEI-3D enables both high-fidelity rendering and fine-grained realistic editing. Fig.~\ref{framework} provides the overall pipeline of our approach. Our overall pipeline consists of three components: \textit{(i)} Given multi-view images with camera parameters, the Explicit-Implicit Reconstruction (EIR) is responsible for reconstructing the object into implicit networks with handler points (Section \ref{ssec: EIR}); \textit{(ii)} Physical Properties Disentangling module (PPD) decouples the color of handler points into learnable parameters and network branches, controlling the physical properties (Section \ref{ssec:SADM}); \textit{(iii)} Spatial-Aware Editing module (SAE) assigns labels for related handler points and realizes part-aware editing. By adjusting the attributes of localized explicit handler points, users can perform fine-grained and realistic edits that drive changes in the implicit surface (Section \ref{ssec:SEM}). Next, we describe each module in detail.

\subsection{Explicit-Implicit Reconstruction}\label{ssec: EIR}
The 3D representation of a target object, reconstructed from multi-view images with camera parameters, directly affects the flexibility and accuracy of the editing process. Existing implicit reconstruction methods utilize a continuous function to represent the target object, which does not support localized editing. Though some methods \cite{yang2022neumesh} combine explicit mesh with implicit network to represent the object and achieve localized editing, they show coarse and inaccurate editing results due to the limited resolution of the mesh. To provide an accurate global and local description of the object, we propose the EIR module,  which constructs a collaborative representation that tightly integrates implicit networks with explicit handler points. The two parts are discussed in detail as follows.

First, we use the Multi-Layer Perceptron (MLP) network to fit the SDF of the target object following \cite{park2019deepsdf, idr, wang2021neus}. The SDF is composed of 8 MLP layers and denoted as $f(\boldsymbol{x};\Phi )$, where $\boldsymbol{x}\in\mathbb{R}^{3}$ denotes the coordinate of a 3D query point and $\Phi $ is the network parameter. 
Then the object surface is located by the zero-level set of SDF, as described in Eq. \eqref{eq:sdf}.
Though Eq. \eqref{eq:sdf} is helpful for object surface modeling, the implicit geometry is stored in the network, i.e., $f(\boldsymbol{x};\Phi)$, making it challenging to be modified locally at edit time. 

To make the implicit representation manipulable, we sample a set of explicit points on the implicit surface, forming a handler points set $\mathcal{H}$. 
The handler points are the intersection points of the implicit surface and the viewing rays, which emanate from the viewpoint of the input image and go through pixels in the image. 
To make the entire pipeline trainable, we sample $\mathcal{H}$ differentiably, expressing them by the parameter $\Phi$ of the network. We represent the intersection point coordinate as $\hat{\boldsymbol{x}}(\Phi)$. Because the intersection point $\hat{\boldsymbol{x}}(\Phi)$ is on both the geometry surface and the ray, it satisfies:
\begin{align}
f(\hat{\boldsymbol{x}}(\Phi);\Phi)=f(\boldsymbol{o}+t(\Phi)\boldsymbol{d};\Phi)=0.
\label{intersecp}
\end{align}
Differentiating Eq. (\ref{intersecp}) w.r.t. $\Phi$, we derive:
\begin{align}
\frac{\partial f(\hat{\boldsymbol{x}})}{\partial \Phi} + \frac{\partial f(\hat{\boldsymbol{x}})}{\partial \hat{\boldsymbol{x}}} \cdot \frac{\partial \hat{\boldsymbol{x}}}{\partial \Phi} = 0.
\label{Differentiating}
\end{align}
We plug the current network parameters $\Phi_{0}$ into the equation to derive the coordinates of the intersection points, which are denoted by $\hat{\boldsymbol{x}}(\Phi_{0})$. We use $\boldsymbol{x}_{0}$, $\boldsymbol{d}_{0}$ to simply represent $\hat{\boldsymbol{x}}(\Phi_{0})$ and corresponding directions and organize Eq. (\ref{Differentiating}) into: 
\begin{align}
    \frac{\partial t}{\partial \Phi}\left(\Phi_{0}\right)=-\frac{1}{\left\langle\frac{\partial f}{\partial \boldsymbol{x}}\left(\boldsymbol{x}_{0} ; \Phi_{0}\right), \boldsymbol{d}_{0}\right\rangle} \frac{\partial f}{\partial \Phi}\left(\boldsymbol{x}_{0} ; \Phi_{0}\right),
    \label{phi}
\end{align}
where $\left\langle\cdot\right\rangle$ denotes the dot product. Now we derive the distance from the camera center as follows: 
\begin{align}
    \label{intersection points}   t\left(\Phi\right)=t_{0}-\frac{1}{\left\langle\frac{\partial f}{\partial \boldsymbol{x}}\left(\boldsymbol{x}_{0} ; \Phi_{0}\right), \boldsymbol{d}_{0}\right\rangle} f\left(\boldsymbol{x}_{0}; \Phi\right). 
\end{align}

Since $f\left(\boldsymbol{x}_{0} ; \Phi_{0}\right)=0$, we get $ t\left(\Phi_{0}\right)=t_{0}$. Plugging Eq. \eqref{intersection points} into $\hat{\boldsymbol{x}}=\boldsymbol{o}+t(\Phi_{0})\boldsymbol{d}$, the intersection point is represented as: 
\begin{align}
    \label{eq: intersection points}
\hat{\boldsymbol{x}}=\mathbf{o}+t_{0}\boldsymbol{d}-\frac{f\left(\boldsymbol{o}+\boldsymbol{t}_{0}\boldsymbol{d};\Phi\right)\boldsymbol{d}}{\left\langle\frac{\partial f}{\partial \boldsymbol{x}}\left(\boldsymbol{x}_{0} ; \Phi_{0}\right), \boldsymbol{d}_{0}\right\rangle}.
\end{align}
The normal $\boldsymbol{n}$ at the point $\hat{\boldsymbol{x}}$ is computed by taking the derivative of the implicit function as:
\begin{align}
\label{eq:normal}
\boldsymbol{n}=\frac{\nabla_{\boldsymbol{x}} f(\hat{\boldsymbol{x}}(\Phi), \Phi)}{\left\|\nabla_{\boldsymbol{x}} f(\hat{\boldsymbol{x}}(\Phi), \Phi)\right\|_{2}},
\end{align}
where $\nabla_{\boldsymbol{x}}$ represents differentiating
w.r.t. $\boldsymbol{x}$.

After deriving the handler points, we represent the target object using the combination of the SDF network and handler point set $\mathcal{H}$. By adjusting the attributes of localized explicit handler points, users can perform fine-grained and controllable edits that drive changes in the implicit surface. Specifically, we determine whether the handler points are to be edited by back-projecting the user's scribbles and independently put the edited points and non-edited points into different branches, ensuring spatial decoupling and preventing the implicit representation from affecting the entire object. When rendering from a new perspective, the appropriate branch is selected based on the distance between the corresponding surface point and the edited handler points, enabling fast and accurate editing. To further enhance editing accuracy, we design a dual-branch network DDA in Section \ref{ssec:SADM} to separately process the edited and non-edited handler points. 

The collaborative representation has the following advantages: \textit{(i)} it makes the implicit representation manipulable with the flexible points (see Fig.~\ref{fig: versatile}) and eliminates the minor artifacts (as shown in Fig.~\ref{fig: two-stage}); \textit{(ii)} it retains the advantages of implicit representations, including high-fidelity rendering and ease of modeling from 2D. Unlike implicit-based editing methods \cite{liu2021editing, wang2023seal}, which inevitably sample points from free space or inside the object, our handler points are sampled directly on the surface, ensuring greater editing accuracy.

\subsection{\mbox{Physical Properties Disentangling}}\label{ssec:SADM}
Based on the reconstructed collaborative 3D representation of the target object in Section \ref{ssec: EIR}, we propose PPD to disentangle environment lighting and material properties of handler points in this section, preparing for realistic editing. The rendered image is physically formed by the combination of lighting, geometry, and material of the object. Therefore, realistic editing requires modification of the specific physical property. However, current advanced implicit 3D editing methods \cite{liu2021editing, yang2022neumesh, xiang2021neutex, Object-Compositional} typically operate directly on the RGB color values, which result in edits that lack realism, giving the impression that the modifications are floating on the object’s surface. In contrast, PPD improves editing realism by disentangling physical properties and editing the specified property.

To achieve editing on the specified property, PPD learns to disentangle lighting parameters and material parameters for handler points.
First, to achieve transform the integral into closed-form expressions, we model the lighting environment map $L_i(\boldsymbol{\omega}_i)$ as the combination of $M$ dimensional Spherical Gaussians (SGs) \cite{some2022sphericalgaussians,learnSphericalGaussians} as follows:
\begin{align}
L_i(\boldsymbol{\omega}_{i}) = \sum_{k=1}^{M} 
\boldsymbol{\mu}_{k} e^{ \lambda_{k} \left ( \boldsymbol{\omega}_{i} \boldsymbol{\xi}_{k}  -1\right)}.
\label{eq: lighting}
\end{align}
The SGs are positioned across the hemisphere centered at $\boldsymbol{n_{\hat{\boldsymbol{x}}}}$. The $\boldsymbol{{\omega}}_{i}\in\mathbb{R}^3$ is the incident lighting direction. $\boldsymbol{\xi}_{k}\in\mathbb{S}^2, \lambda_{k}\in\mathbb{R}_{+},\boldsymbol{\mu}_{k}\in\mathbb{R}_{+}^3$ are learnable parameters, which respectively control the lobe direction, lobe bandwidth, and lobe amplitude of the Gaussian.

To model the diffuse term in rendering equation, we set the metalness as a learnable parameter $\boldsymbol{m}$ and design a network to learn the position-related diffuse albedo $\boldsymbol{a}$. Diffuse albedo describes the base color of the object since it does not contain lighting information, as demonstrated in Eq. (\ref{eq: fd}). Compared with directly modifying the RGB value of a local position on the object surface, modifying the diffuse albedo is more reasonable and makes the user's scribbles realistic. 

To prepare for fine-grained editing, we design a Dual Diffuse Albedo (DDA) network $F_{a}(\gamma(\hat{\boldsymbol{x}}))$ to model the diffuse albedo $\boldsymbol{a}$. The diffuse term is denoted by $\mathcal{F}_{d}$ and is formulated as:
\begin{align}
\mathcal{F}_{d}(\hat{\boldsymbol{x}})=\frac{1-\boldsymbol{m}}{\pi}F_{a}(\gamma(\hat{\boldsymbol{x}})).
\label{eq: our_fd}
\end{align}
In the training phase, we optimize the branch $F_{at}(\gamma(\hat{\boldsymbol{x}});\Psi_{t})$ with training images. In the editing phase, we optimize the branch $F_{ae}(\gamma(\hat{\boldsymbol{x}});\Psi_{e})$ using the scribbled image. For novel-view rendering, the intersection points are routed to either $F_{ae}$ or $F_{at}$ based on their distance to the edited point set denoted by $\mathcal{H}^{+}$. We use the Euclidean distance $\mathcal{D}(\cdot)$ between $\hat{\boldsymbol{x}}$ and the edited handler point set for calculation and introduce a threshold parameter $\theta$ for judgment. If the distance is below the threshold, $\hat{\boldsymbol{x}}$ is classified as an edited position denoted by $\hat{\boldsymbol{x}}_e$ and sent to $F_{ae}$. Otherwise, it is a non-edited position $\hat{\boldsymbol{x}}_u$ and sent to $F_{at}$. Each branch of DDA is composed of a 4-layer MLP with 512 units per layer, referring to the network structures in \cite{nerf, idr}. $\Psi$ is the weight of the MLP. We apply the position encoding $\gamma(\cdot)$ to improve high-frequency details, following NeRF \cite{nerf}.

We model the specular term using the simplified Disney BRDF as used in \cite{bi2020, li2018learning, physg2021} 
and simplify the normalized distribution function using a single SG. The specular term is approximated as:
\begin{align}
\mathcal{F}_s(\boldsymbol{\omega}_o, \boldsymbol{\omega}_i; \hat{\boldsymbol{x}}) = \mathcal{M}_x \boldsymbol{\mu} \cdot e^{\frac{\lambda}{4 (\boldsymbol{h} \cdot \boldsymbol{\omega}_o)} \left((\boldsymbol{h} \cdot \boldsymbol{n}) - 1\right)},
\label{eq: specular}
\end{align}
where $\boldsymbol{h} = \frac{\boldsymbol{\omega}_o + \boldsymbol{\omega}_i}{\|\boldsymbol{\omega}_o + \boldsymbol{\omega}_i\|_2}$. $\mathcal{M}_x$ accounts for the
Fresnel and shadowing effects, which is approximated by a constant calculated as in \cite{wang2009all}, using the roughness $\rho\in\mathcal{R}$, the specular albedo $\alpha$, the incident and view direction and normal at $\hat{\boldsymbol{x}}$. $\lambda\in\mathbb{R}_{+},\boldsymbol{\mu}\in\mathbb{R}_{+}^3$ are learnable parameters, which respectively control lobe bandwidth and lobe amplitude of SG.

After modeling the lighting and reflectance  coefficient of Eq. (\ref{eq: pbr}), we derive color $\hat{\boldsymbol{c}}(\boldsymbol{r})$ as the combination of diffuse color $\hat{\boldsymbol{c}}_d(\boldsymbol{r})$ and specular color $\hat{\boldsymbol{c}}_s(\boldsymbol{r})$, as follows:
\begin{align}
\small
\label{eq: cpred}
\!\!\hat{\boldsymbol{c}}(\boldsymbol{r}) = 
\underset{\hat{\boldsymbol{c}}_d(\boldsymbol{r})}{\underbrace{\sum_{i=1}^{O} L_i(\boldsymbol{\omega}_i) \mathcal{F}_{d}(\hat{\boldsymbol{x}})}} 
+ 
\underset{\hat{\boldsymbol{c}}_s(\boldsymbol{r})}{\underbrace{\sum_{i=1}^{O} L_i(\boldsymbol{\omega}_i) \mathcal{F}_s(\boldsymbol{\omega}_o, \boldsymbol{\omega}_i; \hat{\boldsymbol{x}})}}.
\end{align}

\textbf{Forward Rendering.} To derive the color of a pixel, we first pass the ray $\boldsymbol{r}$ through the pixel to obtain the intersection point $\hat{\boldsymbol{x}}$ and its normal $\boldsymbol{n}$ according to Eq. (\ref{eq: intersection points}) and Eq. (\ref{eq:normal}) in EIR. Then, we derive the diffuse albedo using the dual network $\mathcal{F}_{d}(\hat{\boldsymbol{x}};\Psi)$ and approximate the specular term by Eq. (\ref{eq: specular}) in PPD. Last, using the image loss (Eq.(\ref{colorloss})) in Section \ref{sec: opt}, we optimize the lighting environment parameters $(\boldsymbol{\xi}_{k}, \lambda_{k}, \boldsymbol{\mu}_{k})$, the specular parameters $(\lambda, \boldsymbol{\mu}, \rho, \alpha)$, the diffuse parameters $\Psi$ and the geometry parameters $\Phi$.

\subsection{Spatial-Aware Editing}\label{ssec:SEM} 
Based on the collaborative explicit-implicit representation, we design the Spatial-Aware Editing (SAE) module to support precise and region-aware editing of geometry, material, and lighting. By modifying the attributes of local handler points according to user scribbles, these changes are propagated through the implicit surface representation, allowing consistent updates across the entire object and enabling coherent appearance from novel viewpoints. Existing editing methods \cite{liu2021editing,yang2022neumesh} suffer from limited spatial awareness, requiring labor-intensive manual coloring across multiple views for the modification of one part. For example, coloring the chair arms from the front view may result in the omission of the sides and back of the chair. To address this, we design a Cross-view Propagation-based 3D Segmentation (CPS) strategy, which assigns binary editing labels for handler points. 

\begin{algorithm}[htbp]
\caption{\small CPS strategy.}
\label{algo: cps}
\renewcommand{\algorithmicrequire}{\textbf{Input:}}
\renewcommand{\algorithmicensure}{\textbf{Output:}}
\begin{algorithmic}[1] 
\REQUIRE  Multi-view images $\mathcal{I} = \{ I_k \}_{k=1}^N$ with camera parameters $\mathcal{C} = \{ C_k \}_{k=1}^N$. User scribble $\boldsymbol{E}_{1}$ and the specified negative pixels $\boldsymbol{U}_{1}$.
\ENSURE A set of handler points to be edited, denoted by $\mathcal{H}^+$.
\STATE $\mathcal{H}^+ = []$
\FOR{each image $I_k \in \mathcal{I}$} 
\STATE Generate the segmentation mask $M_{k}$ by SAM model with positive and negative prompt points using $M_{k} = \mathrm{SAM}(I_{k}, \boldsymbol{E}_{k}, \boldsymbol{U}_{k})$;
\STATE Calculate the 3D points that need to be edited by back projection and update the edited points set $\mathcal{H}^+_k = \text{BackProj}(M_{k}, C_{k})$;
\STATE Calculate the positive prompt points for the $(k+1)$-th view by $E_{k+1} = \text{Proj}(\mathcal{H}^+_k, C_{k+1})$;
\STATE Calculate the negative prompt points for the $(k+1)$-th view by $U_{k+1} = \text{Proj}(\mathcal{H}^-_k, C_{k+1})$;
\STATE Add 3D points to be edited into the set $\mathcal{H}^+$ by $\mathcal{H}^+ \gets \mathcal{H}^+ \cup \mathcal{H}^+_k$
\ENDFOR
\STATE \textbf{return} The 3D points set $\mathcal{H}^+$.
\end{algorithmic}
\end{algorithm}

We describe CPS in Algorithm~\ref{algo: cps}. First, a random image is selected from the input images $\mathcal{I}$ and represented as $I_1$ with its camera parameters $C_1$. CPS then allows the user to label the desired editing and non-editing regions using positive and negative 2D prompt point sets $\boldsymbol{E}_1$ and $\boldsymbol{U}_1$, respectively. The CPS derives a binary segmentation mask $M_k$ using the pre-trained segmentation model SAM \cite{sam}. Next, CPS labels the corresponding 3D handler points as edited by back-projecting the desired editing region. Simultaneously, the labeled non-edited 2D points are also back-projected onto the 3D surface and marked as non-edited handler points. To propagate the single scribble to the entire corresponding 3D part and enable spatial-aware editing, CPS projects the labeled edited and non-edited points onto adjacent novel views. These projections serve as prompt points for 2D segmentation using SAM.

This process is repeated across the image set $\mathcal{I}$, allowing CPS to find the handler points to be edited, denoted as $\mathcal{H}^+$. Spatial awareness is achieved by passing the editing signal to all points in $\mathcal{H}^+$ during editing. Note that the part awareness is toggleable. It is enabled for part editing and disabled when drawing precise patterns.

After specifying handler points to be edited, SAE realizes object editing by first modifying the designated physical attributes of these points to the user-specified target values. For fine-grained user scribbles, the updated handler point attributes are used to supervise the corresponding branch (e.g., the diffuse color network $F_{s}$) via the loss function defined in Section \ref{sec: opt}. For geometry editing, the coordinates of handler points are modified first, and then the geometry network is fine-tuned accordingly to adjust the underlying surface geometry. SAE also supports roughness and lighting editing by directly modifying the global roughness or lighting parameters.

\subsection{Model Optimization}
\label{sec: opt}
Following existing editing methods \citep{yang2022neumesh}, the optimization of our model consists of the reconstruction phase and the editing phase. We define the losses as follows. 

\smallskip
\noindent
\textbf{Reconstruction Loss.}
We define the reconstruction loss $\mathcal{L}_r$ as the weighted combination of the color loss and the regularization loss as follows: 
\begin{equation}
\label{colorloss}
\!\mathcal{L}_{r}{=}\frac{1}{N} \sum_{i=1}^{N}\left\|\hat{\boldsymbol{c}}(\boldsymbol{r}){-}\boldsymbol{c}(\boldsymbol{r})\right\|_{1}\!{+}\beta\frac{1}{N_{x}} \sum_{i=1}^{N_{x}}\Big\|\| \nabla_{\boldsymbol{x}_{i}} f\|_{2}{-}1\Big\|_{2}^{2}.\!
\end{equation}

The color loss (first term) ensures the reconstruction quality under the supervision of the ground truth pixel color. $\left\|\cdot\right\|_{1}$ represents the $L_{1}$ norm and $\left\|\cdot\right\|_{2}$ represents the $L_{2}$ norm. $N$ denotes the number of pixels in a mini-batch, $\hat{\boldsymbol{c}}(\boldsymbol{r})$ denotes the rendered pixel color along the ray $\boldsymbol{r}$, and $\boldsymbol{c}(\boldsymbol{r})$ is the ground truth pixel color. The regularization loss (second term) is Eikonal regularization proposed in \citep{gropp2020implicit}, ensuring that surface function $f(\boldsymbol{x}; \Phi)$ satisfies the properties of SDF. The $\left \{\boldsymbol{x}_{i}\right\} _{i=1}^{N_{x}} $ are randomly sampled 3D points inside the bounding box of the object. $\beta$ is the weight balancing the two loss terms.

\noindent
\textbf{Edit Loss.} 
In the editing phase, our model supports scribbling, geometry deformation, and changes in roughness or lighting. For scribbling, we use the scribble color as the target diffuse color $\boldsymbol{c}_{d}(\boldsymbol{r})$. Then we update the diffuse albedo network $F_a$ while freezing the other parameters through the editing loss function $\mathcal{L}_{d}$ as follows:
\begin{equation}
\label{colorloss}
 \mathcal{L}_{d}= \frac{1}{N_{e}} \sum_{i=1}^{N_{e}}\left\|\hat{\boldsymbol{c}}_{d}(\boldsymbol{r})-\boldsymbol{c}_{d}(\boldsymbol{r})\right\|_{1},
\end{equation}
where $\hat{\boldsymbol{c}}_{d}(\boldsymbol{r})$ is the predicted diffuse color and the $N_{e}$ is the number of the edited points.
For geometry editing, we fine-tune the geometry network $f(\boldsymbol{x};\Phi)$ by minimizing the discrepancy between the SDF values predicted by the network and the actual distances to the deformed geometry surface. Denoting the randomly sampled point set as $\{\boldsymbol{s}_i\}_{i=1}^{N_s}$, where $N_s$ is the total number of points. Representing the deformed handler point set as $\mathcal{H}^{'}$, the geometry editing loss $\mathcal{L}_g$ is formulated as: 
\begin{equation}
\label{eq:editcolorloss}
\mathcal{L}_g {=} \frac{1}{N_s} \sum_{i=1}^{N_s} \left\| \min_{\boldsymbol{h} \in \mathcal{H}^{'}} \|\boldsymbol{s}_i - \boldsymbol{h}\|_{1} {-} \boldsymbol{f}(\boldsymbol{s}_i;\Phi) \right\|_{1}.
\end{equation}
For roughness or lighting editing, we modify the corresponding parameters directly without additional optimization. 

\begin{figure*}[ht]
\centering
\includegraphics[width=\linewidth]{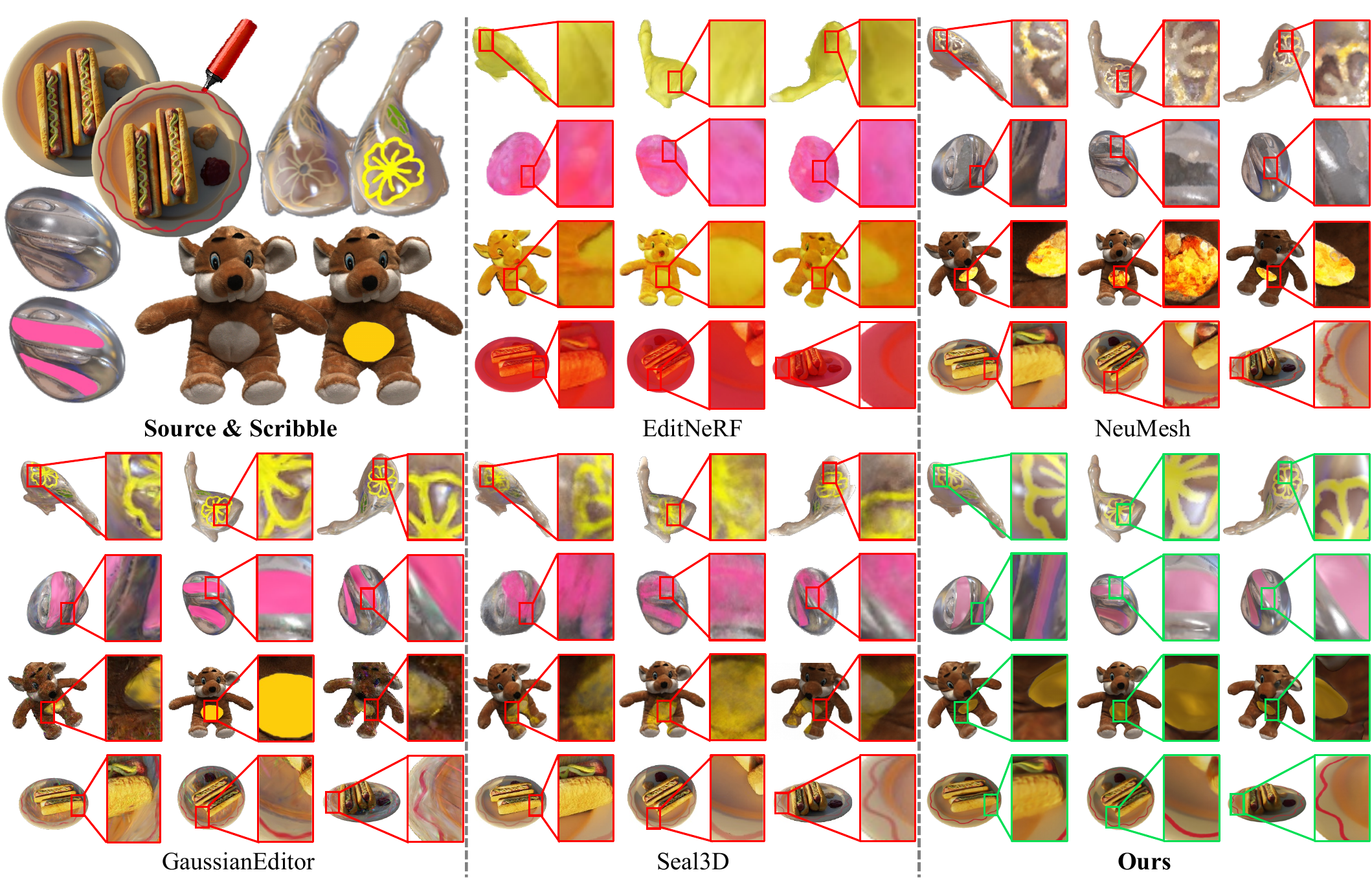}
\caption{\small \textbf{Appearance editing comparisons.} Qualitative comparison of our method with EditNeRF and NeuMesh for pattern painting edits on synthetic and real datasets. From left to right, we show the source image, user scribble, and novel-view rendering of edited results for each method. It can be observed that our approach provides consistent pattern edits on all of the examples.}
\label{fig:texture_editing}
\end{figure*}

\section{Experiments}
\label{experiments}
\subsection{Experimental Setups}
\noindent\textbf{Datasets.} We evaluate our method on four widely used datasets, each with distinct characteristics: the PhySG Synthetic dataset \cite{physg2021} (hereafter referred to as Phy. Syn.), containing objects with glossy surfaces; the NeRF Synthetic dataset \cite{nerf} (hereafter referred to as NeRF Syn.), known for its diverse synthetic scenes; the real-world DTU dataset \cite{DTU}, which offers high-quality 3D scans of real objects; and the chair-class dataset from PhotoShape \cite{PhotoShape}, which includes objects within the same category. For the DTU and PhotoShape datasets, we select 10 objects from each dataset, while for the other two datasets, we use all available data. We follow the official split for Phy. Syn. dataset with resolution $512 \times 512$, utilizing 100 views for training and 100 views for the test. We follow the official split for the NeRF synthetic dataset with resolution $800 \times 800$, utilizing 100 views for training and 200 views for the test. For each scene of the DTU dataset, we randomly split the images of $1600 \times 1200$ resolution into the training set and test set with a ratio of 9:1. The chairs from the PhotoShape dataset are at a resolution of $256 \times 256$. We utilize 40 training and 40 test views for each instance, disregarding 20 validation views. The scribbles we used for testing were provided by four nonprofessionals to ensure that the editing was accessible to ordinary users.

\noindent
\textbf{Evaluation Metrics.} Since the ground truths of the editing results are unavailable, we utilize Frechet Inception Distance (FID)~\cite{FID} as the quantitative metric to evaluate the editing results. The FID metric, which measures the distance between feature distributions of real and generated images, is widely used in image generation tasks \cite{giraffe, graf, tryon}. A lower FID score indicates better similarity between the edited and source images, reflecting higher-quality editing. Additionally, since we aim to preserve the quality of the non-edited area, we introduce image metrics including PSNR (higher is better), SSIM (higher is better), and LPIPS \cite{lpips} (lower is better) to evaluate the reconstruction quality of the non-edited regions.

\noindent
\textbf{Comparison Methods.}
We compare our method qualitatively and quantitatively with four SOTA 3D editing methods: EditNeRF \cite{liu2021editing}, NeuMesh \cite{yang2022neumesh}, Seal3D \cite{wang2023seal}, and GaussianEditor \cite{chen2024gaussianeditor}. EditNeRF and Seal3D primarily focus on editing implicit 3D representations, while NeuMesh and GaussianEditor leverage explicit 3D representations for editing tasks. Since EditNeRF was trained on category-level datasets in the original paper and demonstrates suboptimal performance when trained and tested on general objects. 

\begin{table*}[t]
    \centering
    \caption{\small Quantitative comparison on the non-edited area of novel-view rendering. We report average PSNR (dB)(higher is better)/ SSIM (higher is better)/ LPIPS (lower is better) on each dataset. All the methods are evaluated on unseen test views. Our method achieves the best results on the three metrics, which proves that the local modification performed by our method can better preserve the non-modified area. The best and second-best results are bold and underlined.}
    \small
    \begin{tabular}{p{2.2cm}<{\centering}|p{2.2cm}<{\centering}|p{1.6cm}<{\centering}p{1.6cm}<{\centering}p{1.6cm}<{\centering}p{1.6cm}<{\centering}p{1.6cm}<{\centering}}
        \toprule
        \textbf{Dataset} & \textbf{Methods} & EditNeRF & NeuMesh & Seal3D & GaussianEditor & Ours \\
        \midrule
        \multirow{3}{*}{Phy. Syn.} 
        & PSNR↑ & 19.42& \underline{30.28} &  26.63& 27.42 & \textbf{33.65}  \\
        & SSIM↑ & 0.731& 0.942 &  0.935& \underline{0.965} &  \textbf{0.983} \\
        & LPIPS↓ & 0.324& 0.045 & \underline{0.029} & 0.057 & \textbf{0.022}   \\
        \midrule
        \multirow{3}{*}{NeRF Syn.} 
        & PSNR↑ & 15.31 & \underline{30.94} & 28.32 & 26.89 & \textbf{31.02} \\
        & SSIM↑ & 0.684 & \underline{0.951} & 0.926 & 0.913 & \textbf{0.955} \\
        & LPIPS↓ & 0.552 & \underline{0.043} & 0.075 & 0.102 & \textbf{0.042} \\
        \midrule
        \multirow{3}{*}{PhotoShape} 
        & PSNR↑ & 22.94 & \underline{27.36} & 26.79 & 27.12 & \textbf{29.49} \\
        & SSIM↑ & 0.841 & \underline{0.920} & 0.902 & 0.915 & \textbf{0.972} \\
        & LPIPS↓ & 0.295 & \underline{0.095} & 0.193 & 0.098 & \textbf{0.081} \\
        \midrule
        \multirow{3}{*}{DTU} 
        & PSNR↑ & 20.15 & \underline{28.29} & 26.45 & 24.50 & \textbf{28.63} \\
        & SSIM↑ & 0.756 & \underline{0.921} & 0.896 & 0.882 & \textbf{0.930} \\
        & LPIPS↓ & 0.308 & \underline{0.117} & 0.213 & 0.261 & \textbf{0.112} \\
        \bottomrule
    \end{tabular}
    \label{table: psnr}
\end{table*}

\noindent
\textbf{Implementation Details.}
We build the SDF using an 8-layer MLP and initialize SDF weights $\Phi$ following \cite{sdfini,idr}. We set the Gaussian dimension $M=128$ in the modeling of the lighting environment. We sample incident lights from a uniform distribution, with the number $O=32$. The diffuse albedo is initialized such that the predicted albedo is about 0.5 at all locations inside the object's bounding box. We initialize the lobe sharpness randomly drawn from [95, 125] for the material estimation, while the initial specular albedo is randomly drawn from [0.18, 0.26]. The lobes for the environment map are initialized using a spherical Fibonacci lattice \cite{keinert2015spherical} to ensure uniform distribution on the unit sphere, with monochromatic colors. Weight $\beta$ of the training loss is set to 0.1 and the threshold $\theta$ is set to 0.002 in our experiments. We implemented our approach on PyTorch 1.10 with Python 3.6.3. The experiments were conducted on the NVIDIA GeForce RTX 2080 Ti GPU. We use the Adam optimizer with an initial learning rate of \( 5 \times 10^{-4} \). During the training phase, the model is trained for 200,000 iterations, with the learning rate halved every 25,000 iterations. In the editing phase, we maintain the learning rate at \( 5 \times 10^{-4} \) and train for 500 iterations.

\begin{table*}[ht]
    \small 
    \centering
    \renewcommand{\arraystretch}{1.2} 
    \caption{Quantitative comparison of appearance editing. To quantitatively assess the quality of editing, we use the Frechet Inception Distance (FID) metric between image sets before and after editing, where a lower FID indicates better quality. 
    Our method shows the best result across all datasets. The best and second-best results are bold and underlined.}
    \begin{tabular}{p{2.8cm}<{\centering}|p{2cm}<{\centering} p{2cm}<{\centering} p{2cm}<{\centering} p{2cm}<{\centering} p{2cm}<{\centering}}
        \toprule
        Dataset & EditNeRF & NeuMesh & Seal3D & GaussianEditor & Ours \\
        \midrule
        Phy. Syn. & 270 & \underline{178} & 206 & 235 & \textbf{124} \\
        NeRF Syn. & 196 & 130 & 152 & \underline{145} & \textbf{121} \\
        DTU & 216 & 198 & \underline{186} & 201 & \textbf{175} \\
        PhotoShape & 139 & 131 & \underline{124} & 128 & \textbf{102} \\
        \bottomrule
    \end{tabular}
    \label{tab: fid}
\end{table*}

\begin{table*}[h]
    \small 
    \centering
    \renewcommand{\arraystretch}{1.4} 
    \caption{Efficiency Comparison. We compare the editing time between SOTA methods and our method, demonstrating the advantage of our method in editing efficiency. The best and second-best results are bold and underlined.}
    \begin{tabular}{p{2.8cm}<{\centering}|p{2cm}<{\centering}p{2cm}<{\centering}p{2cm}<{\centering}p{2cm}<{\centering}p{2cm}<{\centering}}
        \toprule
        Dataset & EditNeRF & NeuMesh & Seal3D & GaussianEditor & Ours \\
        \midrule
        Phy. Syn. & 37s & 1.2h & \underline{29s} & 44s & \textbf{14s} \\
        NeRF Syn. & \underline{28s} & 1.3h & 32s & 38s & \textbf{18s} \\
        DTU & 76s & 1.6h & \underline{34s} & 42s & \textbf{20s} \\
        PhotoShape & 19s & 1.1h & \underline{28s} & 36s & \textbf{12s} \\
        \hline
        Mean & 48s & 1.4h & \underline{32s} & 40s & \textbf{16s} \\
        \bottomrule
    \end{tabular}
    \label{tab: efficiency}
\end{table*}

\subsection{Comparisons}
We present a comprehensive comparison between our method and several recent state-of-the-art (SOTA) approaches across three dimensions: visual results, quantitative evaluation, and editing efficiency. We focus our analysis on two key aspects of editing quality: \textit{(i)} editing accuracy, which ensures that the edited elements stay true to the user's scribble, with clear boundaries, correct patterns, and consistent colors that faithfully reflect the user's intent while preserving the integrity of the non-edited regions; \textit{(ii)} editing realism, which involves the seamless blending of the edited elements into the environment, ensuring they integrate naturally and appear convincingly realistic within the scene.

\noindent
\textbf{Visual Comparison.} 
The visual results are shown in Fig.~\ref{fig:texture_editing}. Our method achieves the best editing results, as evidenced by clear and accurate editing patterns across various viewpoints. The scribbles are seamlessly integrated into the lighting, and the non-edited areas remain faithful to the source object without being affected by the editing. The method that performs second best to ours is NeuMesh, which generally produces edits that align with the user’s scribble. However, NeuMesh exhibits some inaccuracies in the pattern color and edges. The mistakes made by NeuMesh primarily stem from its method of obtaining colors through the interpolation of mesh vertices. This approach inadvertently incorporates colors from non-edited areas, leading to errors. Additionally, the resolution of the mesh limits the level of detail that can be edited, further exacerbating these inaccuracies. Due to the spatial coupling of EditNeRF, the edited color spills over the whole object, resulting in the most severe undesirable color changes. Similar undesired color changes are shown in Seal3D, which is also a continuous NeRF-based editing method. Though GaussianEditor uses a discrete Gaussian representation, it exhibits unwanted Gaussian changes, which are shown as thorns and bright spots in the images. Besides, the results of all SOTA methods lack realism, more like a pattern on the image rather than truly integrating into the environment to change the object itself. 

In contrast, our method benefits from the collaborative explicit-implicit representation, where localized edits are precisely applied to explicit handler points, and the implicit surface ensures smooth and consistent propagation across views. This synergy enables natural integration of user edits into the object's appearance under complex lighting conditions. The benefits of our realistic editing are particularly evident on shiny objects. As shown in the first two rows of Fig.~\ref{fig:texture_editing}, even though the original mouse is shiny, we handle reflections well and naturally blend the pink color into the mouse.

\noindent
\textbf{Quantitative Comparison.}
The quantitative comparison is conducted from three key perspectives: \textit{(i)} the editing realism evaluated by FID, which is quantified by the similarity of the image sets before and after editing; \textit{(ii)} the editing accuracy, which is evaluated by the quality of the non-edited region using PSNR, SSIM, and LPIPS, to ensure that these areas are preserved with high fidelity and remain free from unintended alterations; \textit{(iii)} the inclusion of a user study to evaluate the editing results, further enriching the evaluation by incorporating subjective judgments. Together, these metrics provide a comprehensive analysis of both the accuracy of the edited regions and the preservation of the original content in the non-edited areas, ensuring that the editing process maintains both realism and high fidelity.

First, we show the FID results in Table~\ref{tab: fid}, which provides a quantitative comparison of appearance editing quality across several datasets. Our method outperforms all other methods, achieving the lowest FID score on each dataset. Notably, on the ``Phy. Syn." dataset that consists of shiny objects, our method achieves an FID of 124, which is 30\% lower than NeuMesh and 47\% lower than EditNeRF, demonstrating a significant improvement in editing realism. This improvement is because our method separates the attributes of the object from the environment, allowing the edits to naturally integrate with the surrounding lighting. This makes the edited images appear more realistic and similar to real-world images. Even on the real-world DTU dataset, our method achieves an FID of 175, outperforming the second-best method Seal3D by 6\%, further demonstrating the superior effectiveness of our approach across various datasets.

Second, we show novel-view rendering results of non-edited regions in Table~\ref{table: psnr}. Our method outperforms all other methods on all four datasets, achieving the highest PSNR, SSIM, and the lowest LPIPS scores. This result is consistent with the visual performance, where our method excels in fine-grained editing without affecting the non-edited areas. In contrast, EditNeRF, Seal3D, and GaussianEditor exhibit unintended alterations in the non-edited regions, caused by varying levels of spatial coupling. On the ``Phy. Syn." dataset, which features shiny objects with obvious lighting, our method achieves a PSNR of 33.65dB, which is 3.37dB higher than the second-best method, NeuMesh (30.28dB), achieving 11.1\% improvement. The quantitative advanced results are specifically reflected in the fact that our renderings effectively retain the high-frequency information in the image, making the details of the objects look clear and sharp.

\begin{figure}[htbp]
\includegraphics[width=1.0\linewidth]{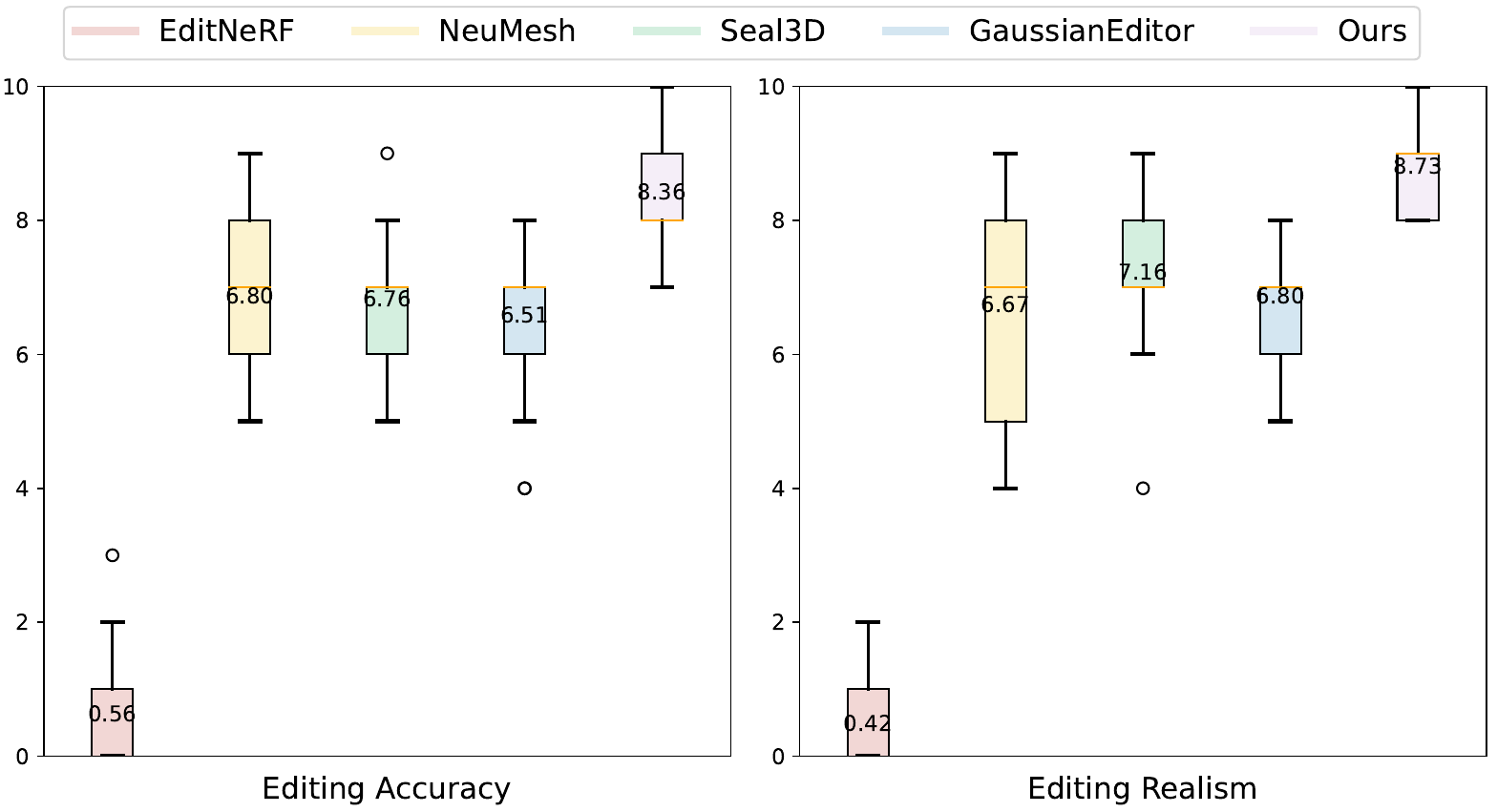}
\caption{\small Boxplot illustration of user study. Our method
demonstrates superior performance (high means) and greater stability across test objects (narrow interquartile range) in both editing accuracy and realism.}
\label{fig: user}
\end{figure}

Lastly, while the above metrics provide valuable insights, they do not fully capture the accuracy and realism of the edits. To address this, we introduce user evaluation to further assess the effectiveness of our method. We conducted user studies with 45 participants to evaluate different methods based on user preferences across 10 randomly selected objects. Participants are asked to assign a preference score (integer ranging from 1 to 10) for editing accuracy and editing realism, based on five random views per scene from the results of the anonymized method. As shown in Fig.~\ref{fig: user}, we report the distribution of the scores, including medians, means, quartiles, and outliers. Our results show that our method is significantly preferred over all baselines in terms of editing accuracy (mean = 8.36) and editing realism (mean = 8.73). The narrow interquartile range of our method also demonstrates a more robust editing performance across various objects.

\noindent
\textbf{Efficiency Comparison.}
Editing time is an important factor that affects the user editing experience. Therefore, we compare the editing time of different methods in Table~\ref{tab: efficiency}. Our method offers a significant advantage in editing efficiency, achieving an average editing time of 16 seconds. This makes our method 315 times faster than NeuMesh (which exhibits the second-best performance in overall editing quality), and nearly 2 times faster than Seal3D, the second-fastest method (which takes 32 seconds). 

\subsection{Ablation Study}\label{ssec:ablation}
In this section, we evaluate the effectiveness of the proposed components by designing and testing different variants based on our overall pipeline. 

\begin{figure*}[t]
\centering
\includegraphics[width=1.0\linewidth]{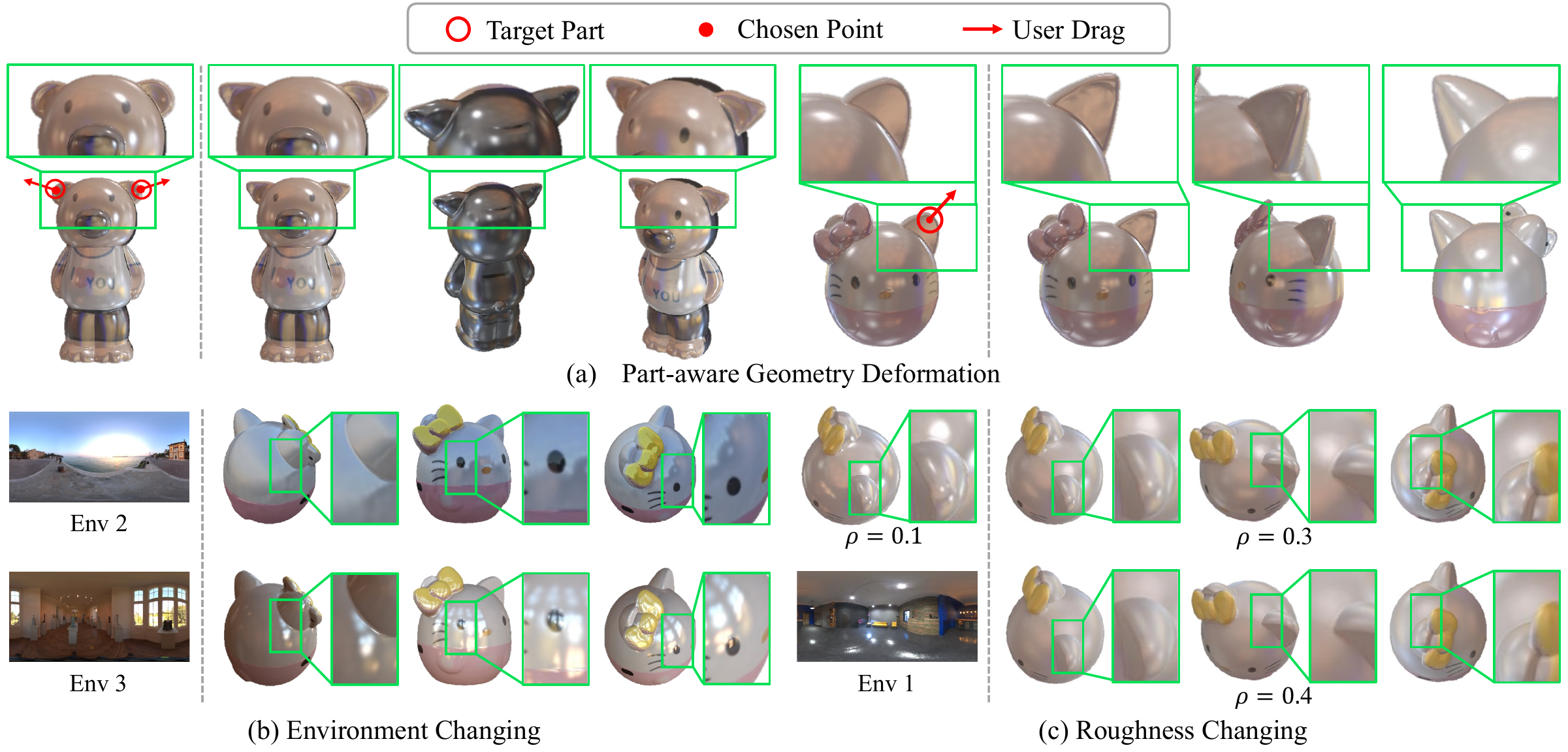}
\caption{\small More editing applications. In addition to the appearance editing, our framework supports part-aware geometry deformation, part removal, environment lighting changing and material roughness editing.}
\label{fig: versatile}
\end{figure*}

\begin{figure}[ht]
\centering
\includegraphics[width=1\linewidth]{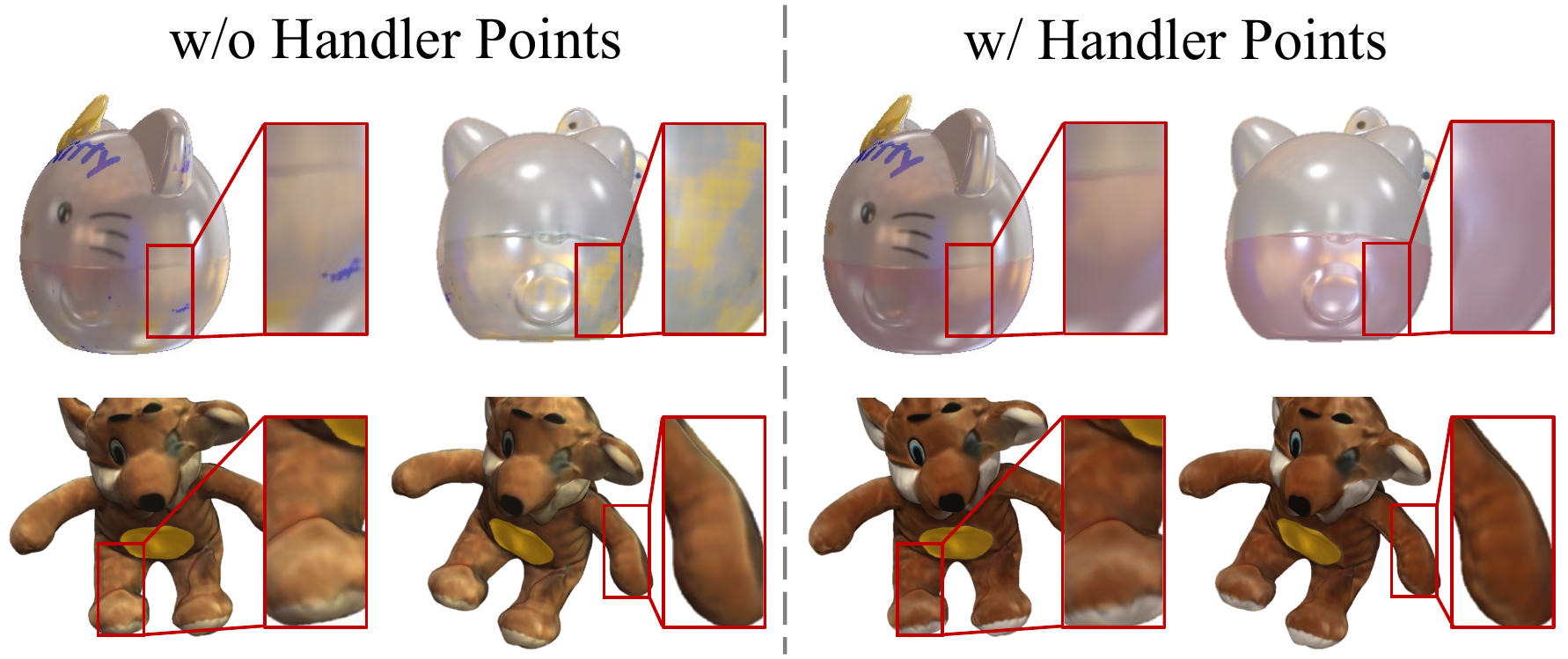}
\caption{\small Effect of handler points. The artifacts annotated in the red box on the left side are eliminated with the handler points.}
\label{fig: two-stage}
\end{figure}
\begin{figure}[h]
\centering
\includegraphics[width=1\linewidth]{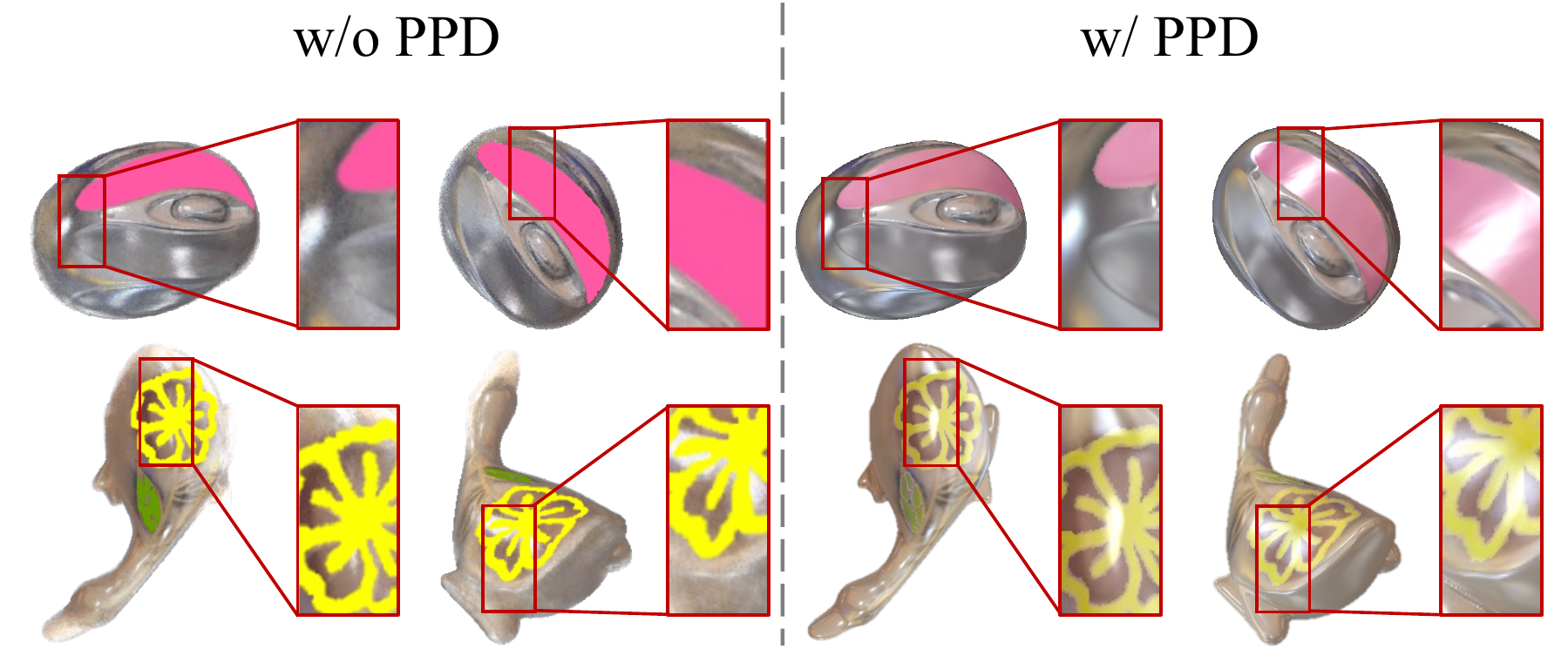}
\caption{\small Effect of PPD. Properties disentangling ensures that the scribbles blend seamlessly into the environment and the editing results are realistic.}
\label{fig:ablation-properties disentangling}
\end{figure}

\noindent
\textbf{Effect of Handler Points.}
We evaluate the effectiveness of handler points in our EIR module by removing them to establish a baseline. As shown in Fig.~\ref{fig: two-stage}, the inclusion of the handler points set significantly eliminates artifacts and undesired color changes in non-edited regions, thereby ensuring accurate editing and high-fidelity rendering. 
Without explicit handler points, editing relies solely on the implicit field, which lacks localized control and makes it difficult to constrain modifications to user-intended areas. This often results in spatial leakage, such as color spillover and inconsistent updates across views. In contrast, our collaborative explicit-implicit design leverages handler points as editable anchors that precisely capture user intent and constrain the influence region of edits. These explicit modifications are then propagated through the implicit surface to ensure smooth, view-consistent updates.

The collaboration between discrete explicit representation and implicit continuous representation enables our method to effectively combine the strengths of both paradigms. This allows our proposed CEI-3D to achieve localized editing and global high-fidelity rendering.

\noindent
\textbf{Effect of PPD.} To assess the impact of properties disentangling in the PPD module, we design a variant with coupled properties as a baseline, where the RGB values of points are optimized directly. As shown in Fig.~\ref{fig:ablation-properties disentangling}, our disentangling approach demonstrates two key advantages over the coupled reconstruction method: \textit{(i)} properties disentangling enhances reconstruction quality, particularly in capturing high-frequency shiny details on the object; and \textit{(ii)} it enables more realistic and physically plausible editing, as user scribbles are integrated into the scene in a manner consistent with the source illumination, rather than appearing artificially overlaid.

\noindent
\textbf{Effect of CPS.} We demonstrate the effectiveness of CPS strategy in the SAE module through a case study where part information simplifies user editing operations. Without CPS, a local scribble cannot modify the entire object part. For instance, if a user wants to change the color of a chair's backrest, they must manually color each pixel of the backrest across multiple views. CPS alleviates this burden by leveraging cross-view consistency to propagate 2D scribbles onto 3D handler points, enabling automatic part labeling and region-aware editing. Integrated into our collaborative explicit-implicit framework, CPS further realizes spatial-aware editing on the continuous surface and makes 3D editing easier and more efficient.

\begin{figure*}[t]
\centering
\includegraphics[width=1\linewidth]{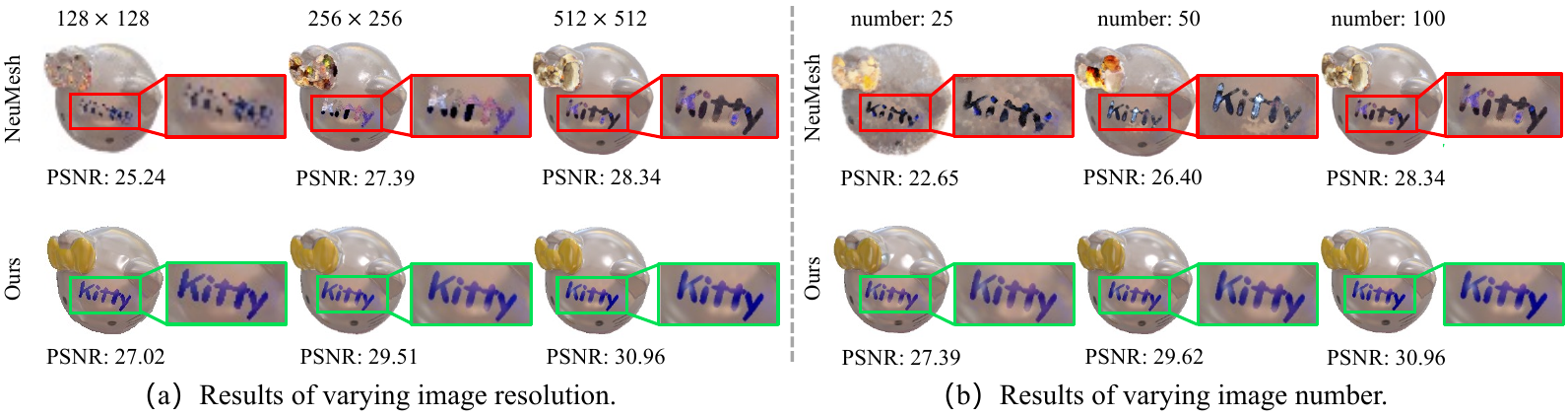}\vspace{-0.1cm}
\caption{\small Robustness Analysis. (a) Editing results of inputs with different resolutions.(b) Editing results of different input image numbers. Our approach shows stronger robustness than the comparison method on both rendering fidelity and editing quality. We use the same scribble as Fig. \ref{fig:intro-img}. PSNR is reported in dB.}
\label{fig: robustness}
\end{figure*}

\begin{figure}[h]
\centering
\includegraphics[width=1\linewidth]{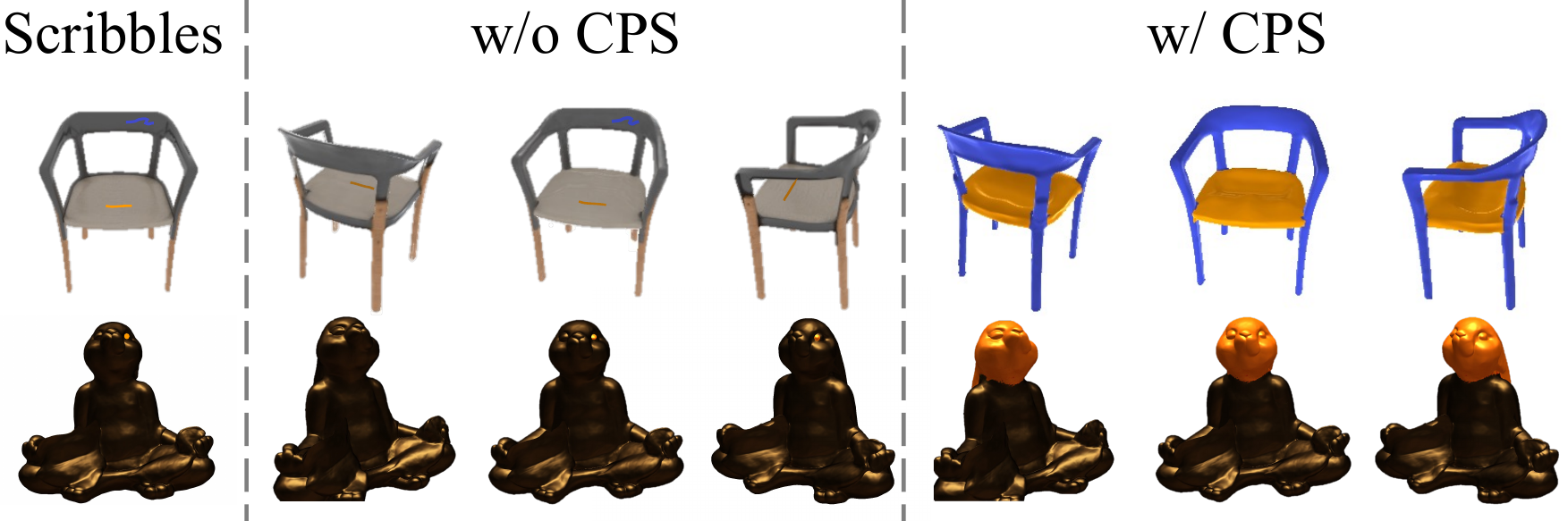}
\caption{\small Effect of CPS. The CPS strategy enables part-aware editing with sparse and single-view scribbles, making the process more efficient and intuitive.}
\end{figure}
\subsection{More Applications}
In addition to appearance modifications, our method can also be flexibly applied to a variety of editing tasks, including geometry editing, relighting and material changes. This helps ordinary users easily complete their creative tasks. 

\noindent
\textbf{Part-aware Geometry Editing.} 
We present part deformation results in Fig.~\ref{fig: versatile}(a). Users specify a geometric part of the target object by adding positive and negative points to the 2D images. Once our method identifies the manipulation area, users drag a handler point within this region to a new position. Our approach uses ARAP \cite{arap} to deform the specified part by displacing handler points in the manipulation area. After the displacement, we fine-tune the geometry network branch using the updated handler points, while fixing other branches, ensuring seamless geometric editing and maintaining a coherent appearance. 

\noindent
\textbf{Environment Changing.} 
Fig.~\ref{fig: versatile}(b) showcases an edited kitty with yellow butter rendered under different lighting environments (Env1, Env2, and Env3). This capability allows for flexible adaptation to various scenarios, making our approach ideal for applications such as augmented reality and creative content generation.

\noindent
\textbf{Roughness Changing.} 
By disentangling object attributes, our model allows modification of specific properties. We adjust the roughness (denoted as $\rho$) of the kitty and present the results in Fig.~\ref{fig: versatile}(c). As the roughness increases, the kitty’s surface becomes less reflective and exhibits a more diffuse appearance, resulting in a softer, less shiny look. This demonstrates the model's ability to effectively control material parameters, enabling realistic customization for diverse visual effects and applications.

\noindent
\subsection{Robustness Analysis} We evaluate the robustness of our method by varying the number and resolution of input images. We compare the results with NeuMesh (the best-performing SOTA method) and present the visual and quantitative results in Fig.~\ref{fig: robustness}. We evaluate the editing capability of the models using the same user-provided scribble (painting the entire butter yellow), to focus on the editing robustness. 

First, we conduct experiments on images with varying resolutions, as shown in Fig.~\ref{fig: robustness} (a). As the resolution of the input image decreases, the user's scribbles in Neumesh's results become progressively less clear, with more noticeable errors and gaps. In contrast, the editing results of our method consistently maintain acceptable clarity.

Next, we fix the resolution at $512 \times 512$ and train both our model and NeuMesh using varying numbers of input images. As shown in Fig.~\ref{fig: robustness} (b), both methods show improvement as the number of input images increases. However, our method consistently outperforms NeuMesh in both rendering and editing quality. Our method maintains stable editing performance for different numbers. In contrast, NeuMesh experiences a gradual decline in editing quality as fewer images are provided.

Overall, our method demonstrates robust editing performance under varying input conditions, excelling in both accuracy and naturalness.

\section{Conclusion}
\label{conclusion}
We propose CEI-3D, an editing-oriented 3D reconstruction framework that enables realistic and fine-grained object editing by reconstructing the 3D object in an inherently editable form. CEI-3D establishes a collaborative explicit-implicit representation that combines the high-fidelity implicit neural fields with the editable and flexible explicit handler points. By adjusting the attributes of these handler points, users can perform precise edits that propagate through the implicit surface, ensuring global consistency. Building on this collaborative structure, CEI-3D disentangles physical properties, enabling independent control over specified properties to realize realistic editing results. Additionally, we incorporate a spatial-aware editing module that leverages cross-view segmentation to support part-aware editing using minimal user input, further enhancing editing efficiency and intuitiveness. Looking ahead, we plan to extend this framework to dynamic scenes and videos, exploring its potential in practical applications such as AR/VR content creation, digital twins, and interactive design. We believe the collaborative explicit-implicit representation will play a vital role for controllable and high-quality 3D editing in the future.

\bmhead{Acknowledgements}{
This work was supported by the National Natural Science Foundation of China (62431015).}

\bmhead{Data Availability}{
All datasets used in this work are publicly available at the following URLs:

\begin{itemize}
  \item Phy. Syn. dataset \citep{physg2021}: \href{https://drive.google.com/drive/folders/1WkecJrB0wUcYYB_T399KCVfc16e3qNLs?usp=drive_link}{[Google Drive Link]}
  \item NeRF Syn. dataset \citep{nerf}: \href{https://drive.google.com/file/d/1OsiBs2udl32-1CqTXCitmov4NQCYdA9g/view?usp=drive_link}{[Google Drive Link]}
  \item DTU dataset \citep{DTU}: \href{https://roboimagedata.compute.dtu.dk/?page_id=36}{[Official Website]}
  \item PhotoShape dataset \citep{PhotoShape}: \href{https://photoshape.github.io/}{[Official Website]}
\end{itemize}}

\bibliographystyle{sn-mathphys-num}
\bibliography{sn-bibliography}

@STRING{AAAI     = "AAAI Conference on Artificial Intelligence"}

@STRING{ACM      = "Association for Computing Machinery"}

@STRING{ARXIV    = "arXiv Preprint"}

@STRING{CACM     = "Communications of the ACM"}

@string{CHI      = "ACM Conference on Human Factors in Computing Systems"}

@STRING{ECCV     = "European Conference on Computer Vision"}

@STRING{NIPS     = "Advances in Neural Information Processing Systems"}

@STRING{SIGGRAPH = "ACM Siggraph Computer Graphics"}

@STRING{TOG      = "ACM Transactions on Graphics"}

@inproceedings{physg2021,
title={PhySG: Inverse Rendering with Spherical Gaussians for Physics-based Material Editing and Relighting},
author={Kai Zhang and Fujun Luan and Qianqian Wang and Kavita Bala and Noah Snavely},
booktitle = {Proceedings of the IEEE/CVF Conference on Computer Vision and
Pattern Recognition},
year={2021}}

@inproceedings{liu2021editing,
  title={Editing Conditional Radiance Fields},
  author={Steven Liu and Xiuming Zhang and Zhoutong Zhang and Richard Zhang and Jun-Yan Zhu and Bryan Russell},
  booktitle = {Proceedings of the IEEE/CVF International Conference on Computer Vision},
  year={2021}
}

@article{PhotoShape,
author = {Park, Keunhong and Rematas, Konstantinos and Farhadi, Ali and Seitz, Steven M.},
title = {PhotoShape: Photorealistic Materials for Large-Scale Shape Collections},
year = {2018},
volume = {37},
number = {6},
journal={ACM Transactions on Graphics},
numpages = {12},
}

@inproceedings{FID,
  title={Gans trained by a two time-scale update rule converge to a local nash equilibrium},
  author={Heusel, Martin and Ramsauer, Hubert and Unterthiner, Thomas and Nessler, Bernhard and Hochreiter, Sepp},
  booktitle =NIPS,
  volume={30},
  year={2017}
}

@inproceedings{graf,
  title={Graf: Generative radiance fields for 3d-aware image synthesis},
  author={Schwarz, Katja and Liao, Yiyi and Niemeyer, Michael and Geiger, Andreas},
  booktitle =NIPS,
  volume={33},
  year={2020}
}

@InProceedings{giraffe,
    author    = {Niemeyer, Michael and Geiger, Andreas},
    title     = {GIRAFFE: Representing Scenes As Compositional Generative Neural Feature Fields},
    booktitle = {Proceedings of the IEEE/CVF Conference on Computer Vision and Pattern Recognition},
    year      = {2021}
}

@article{CCNERF,
  title={Compressible-composable NeRF via Rank-residual Decomposition},
  author={Tang, Jiaxiang and Chen, Xiaokang and Wang, Jingbo and Zeng, Gang},
  journal={arXiv preprint arXiv:2205.14870},
  year={2022}
}

@article{nerf,
  title={Nerf: Representing scenes as neural radiance fields for view synthesis},
  author={Mildenhall, Ben and Srinivasan, Pratul P and Tancik, Matthew and Barron, Jonathan T and Ramamoorthi, Ravi and Ng, Ren},
  journal=CACM,
  volume={65},
  number={1},
  pages={99--106},
  year={2021}
}

@inproceedings{idr,
  title={Multiview Neural Surface Reconstruction by Disentangling Geometry and Appearance},
  author={Yariv, Lior and Kasten, Yoni and Moran, Dror 
      and Galun, Meirav and Atzmon, Matan and Ronen, Basri and Lipman, Yaron},
  booktitle =NIPS,
  volume={33},
  year={2020}
}

@article{zhang2021nerfactor,
  title={Nerfactor: Neural factorization of shape and reflectance under an unknown illumination},
  author={Zhang, Xiuming and Srinivasan, Pratul P and Deng, Boyang and Debevec, Paul and Freeman, William T and Barron, Jonathan T},
  journal=TOG,
  volume={40},
  number={6},
  pages={1--18},
  year={2021}
}

@inproceedings{mescheder2019occupancy,
  title={Occupancy networks: Learning 3d reconstruction in function space},
  author={Mescheder, Lars and Oechsle, Michael and Niemeyer, Michael and Nowozin, Sebastian and Geiger, Andreas},
  booktitle={Proceedings of the IEEE/CVF Conference on Computer Vision and Pattern Recognition},
  year={2019}
}

@inproceedings{lorensen1998marching,
  title={Marching cubes: A high resolution 3D surface construction algorithm},
  author={Lorensen, William E and Cline, Harvey E},
  booktitle=SIGGRAPH,
  pages={347--353},
  year={1998}
}

@inproceedings{p2m++,
  title={Pixel2mesh++: Multi-view 3d mesh generation via deformation},
  author={Wen, Chao and Zhang, Yinda and Li, Zhuwen and Fu, Yanwei},
  booktitle = {Proceedings of the IEEE/CVF International Conference on Computer Vision},
  pages={1042--1051},
  year={2019}
}

@inproceedings{sdfini,
  title={Sal: Sign agnostic learning of shapes from raw data},
  author={Atzmon, Matan and Lipman, Yaron},
  booktitle={Proceedings of the IEEE/CVF Conference on Computer Vision and Pattern Recognition},
  year={2020}
}

@inproceedings{park2019deepsdf,
  title={Deepsdf: Learning continuous signed distance functions for shape representation},
  author={Park, Jeong Joon and Florence, Peter and Straub, Julian and Newcombe, Richard and Lovegrove, Steven},
  booktitle={Proceedings of the IEEE/CVF Conference on Computer Vision and Pattern Recognition},
  year={2019}
}

@inproceedings{tryon,
  title={Parser-free virtual try-on via distilling appearance flows},
  author={Ge, Yuying and Song, Yibing and Zhang, Ruimao and Ge, Chongjian and Liu, Wei and Luo, Ping},
  booktitle={Proceedings of the IEEE/CVF Conference on Computer Vision and Pattern Recognition},
  year={2021}
}

@inproceedings{gropp2020implicit,
    author = {Gropp, Amos and Yariv, Lior and Haim, Niv and Atzmon, Matan and Lipman, Yaron},
    title = {Implicit geometric regularization for learning shapes},
    year = {2020},
    booktitle = {International Conference on Machine Learning},
    articleno = {355},
    numpages = {11}
}

@inproceedings{DTU,
  title={Large scale multi-view stereopsis evaluation},
  author={Jensen, Rasmus and Dahl, Anders and Vogiatzis, George and Tola, Engin and Aan{\ae}s, Henrik},
  booktitle={Proceedings of the IEEE/CVF Conference on Computer Vision and Pattern Recognition},
  year={2014}
}

@inproceedings{yang2022neumesh,
  title={Neumesh: Learning disentangled neural mesh-based implicit field for geometry and texture editing},
  author={Yang, Bangbang and Bao, Chong and Zeng, Junyi and Bao, Hujun and Zhang, Yinda and Cui, Zhaopeng and Zhang, Guofeng},
  booktitle=ECCV,
  year={2022}
}

@inproceedings{wangibrnet,
  author    = {Wang, Qianqian and Wang, Zhicheng and Genova, Kyle and Srinivasan, Pratul and Zhou, Howard  and Barron, Jonathan T. and Martin-Brualla, Ricardo and Snavely, Noah and Funkhouser, Thomas},
  title     = {IBRNet: Learning Multi-View Image-Based Rendering},
  booktitle={Proceedings of the IEEE/CVF Conference on Computer Vision and Pattern Recognition},
  year      = {2021}
}

@inproceedings{arap,
  title={As-rigid-as-possible surface modeling},
  author={Sorkine, Olga and Alexa, Marc},
  booktitle={Symposium on Geometry processing},
  volume={4},
  pages={109--116},
  year={2007}
}

@InProceedings{Object-Compositional,
    author    = {Yang, Bangbang and Zhang, Yinda and Xu, Yinghao and Li, Yijin and Zhou, Han and Bao, Hujun and Zhang, Guofeng and Cui, Zhaopeng},
    title     = {Learning Object-Compositional Neural Radiance Field for Editable Scene Rendering},
    booktitle = {Proceedings of the IEEE/CVF International Conference on Computer Vision},
    month     = {October},
    year      = {2021},
}

@inproceedings{xiang2021neutex,
  title={Neutex: Neural texture mapping for volumetric neural rendering},
  author={Xiang, Fanbo and Xu, Zexiang and Hasan, Milos and Hold-Geoffroy, Yannick and Sunkavalli, Kalyan and Su, Hao},
  booktitle={Proceedings of the IEEE/CVF Conference on Computer Vision and Pattern Recognition},
  year={2021}
}

@inproceedings{tancik2022block,
  title={Block-nerf: Scalable large scene neural view synthesis},
  author={Tancik, Matthew and Casser, Vincent and Yan, Xinchen and Pradhan, Sabeek and Mildenhall, Ben and Srinivasan, Pratul P and Barron, Jonathan T and Kretzschmar, Henrik},
  booktitle={Proceedings of the IEEE/CVF Conference on Computer Vision and Pattern Recognition},
  pages={8248--8258},
  year={2022}
}

@inproceedings{xie2021fig,
  title={Fig-nerf: Figure-ground neural radiance fields for 3d object category modelling},
  author={Xie, Christopher and Park, Keunhong and Martin-Brualla, Ricardo and Brown, Matthew},
  booktitle={International Conference on 3D Vision},
  pages={962--971},
  year={2021}
}

@inproceedings{NeRF-editing,
  title={NeRF-editing: geometry editing of neural radiance fields},
  author={Yuan, Yu-Jie and Sun, Yang-Tian and Lai, Yu-Kun and Ma, Yuewen and Jia, Rongfei and Gao, Lin},
  booktitle={Proceedings of the IEEE/CVF Conference on Computer Vision and Pattern Recognition},
  pages={18353--18364},
  year={2022}
}

@article{muller2022instant,
  title={Instant neural graphics primitives with a multiresolution hash encoding},
  author={M{\"u}ller, Thomas and Evans, Alex and Schied, Christoph and Keller, Alexander},
  journal={ACM Transactions on Graphics},
  volume={41},
  number={4},
  pages={1--15},
  year={2022}
}

@article{kerbl20233dgs,
  title={3D Gaussian Splatting for Real-Time Radiance Field Rendering},
  author={Kerbl, Bernhard and Kopanas, Georgios and Leimk{\"u}hler, Thomas and Drettakis, George},
  journal={ACM Transactions on Graphics},
  volume={42},
  number={4},
  year={2023}
}

@misc{zbrush2023,
  author = {Pixologic},
  title = {ZBrush},
  howpublished = {\url{https://www.pixologic.com}},
  year = {2023}
}

@inproceedings{lee2023ice,
  title={Ice-nerf: Interactive color editing of nerfs via decomposition-aware weight optimization},
  author={Lee, Jae-Hyeok and Kim, Dae-Shik},
  booktitle={Proceedings of the IEEE/CVF International Conference on Computer Vision},
  pages={3491--3501},
  year={2023}
}

@article{li2022language,
  title={Language-driven semantic segmentation},
  author={Li, Boyi and Weinberger, Kilian Q and Belongie, Serge and Koltun, Vladlen and Ranftl, Ren{\'e}},
  journal={arXiv preprint arXiv:2201.03546},
  year={2022}
}

@inproceedings{kuang2023palettenerf,
  title={Palettenerf: Palette-based appearance editing of neural radiance fields},
  author={Kuang, Zhengfei and Luan, Fujun and Bi, Sai and Shu, Zhixin and Wetzstein, Gordon and Sunkavalli, Kalyan},
  booktitle={Proceedings of the IEEE/CVF Conference on Computer Vision and Pattern Recognition},
  pages={20691--20700},
  year={2023}
}

@inproceedings{disneybrdf,
  title={Physically-based shading at disney},
  author={Burley, Brent and Studios, Walt Disney Animation},
  booktitle=SIGGRAPH,
  year={2012}
}

@inproceedings{sam,
  title={Segment anything},
  author={Kirillov, Alexander and Mintun, Eric and Ravi, Nikhila and Mao, Hanzi and Rolland, Chloe and Gustafson, Laura and Xiao, Tete and Whitehead, Spencer and Berg, Alexander C and Lo, Wan-Yen and others},
  booktitle={Proceedings of the IEEE/CVF International Conference on Computer Vision},
  pages={4015--4026},
  year={2023}
}

@article{xu2024instantmesh,
  title={Instantmesh: Efficient 3d mesh generation from a single image with sparse-view large reconstruction models},
  author={Xu, Jiale and Cheng, Weihao and Gao, Yiming and Wang, Xintao and Gao, Shenghua and Shan, Ying},
  journal={arXiv preprint arXiv:2404.07191},
  year={2024}
}

@article{wang2021neus,
  title={Neus: Learning neural implicit surfaces by volume rendering for multi-view reconstruction},
  author={Wang, Peng and Liu, Lingjie and Liu, Yuan and Theobalt, Christian and Komura, Taku and Wang, Wenping},
  journal={arXiv preprint arXiv:2106.10689},
  year={2021}
}

@article{samavati2023survey,
  title={Deep learning-based 3D reconstruction: a survey},
  author={Samavati, Taha and Soryani, Mohsen},
  journal={Artificial Intelligence Review},
  volume={56},
  number={9},
  pages={9175--9219},
  year={2023}
}

@inproceedings{wang2023neus2,
  title={Neus2: Fast learning of neural implicit surfaces for multi-view reconstruction},
  author={Wang, Yiming and Han, Qin and Habermann, Marc and Daniilidis, Kostas and Theobalt, Christian and Liu, Lingjie},
  booktitle={Proceedings of the IEEE/CVF International Conference on Computer Vision},
  pages={3295--3306},
  year={2023}
}

@article{schirmer2024surveyofsdf,
  title={Geometric implicit neural representations for signed distance functions},
  author={Schirmer, Luiz and Novello, Tiago and da Silva, Vin{\'\i}cius and Schardong, Guilherme and Perazzo, Daniel and Lopes, H{\'e}lio and Gon{\c{c}}alves, Nuno and Velho, Luiz},
  journal={Computers \& Graphics},
  pages={104085},
  year={2024}
}

@inproceedings{chen2024gaussianeditor,
  title={Gaussianeditor: Swift and controllable 3d editing with gaussian splatting},
  author={Chen, Yiwen and Chen, Zilong and Zhang, Chi and Wang, Feng and Yang, Xiaofeng and Wang, Yikai and Cai, Zhongang and Yang, Lei and Liu, Huaping and Lin, Guosheng},
  booktitle={Proceedings of the IEEE/CVF Conference on Computer Vision and Pattern Recognition},
  pages={21476--21485},
  year={2024}
}

@inproceedings{zhu2023i2sdf,
    title={I2-SDF: Intrinsic Indoor Scene Reconstruction and Editing via Raytracing in Neural SDFs},
    author={Zhu, Jingsen and Huo, Yuchi and Ye, Qi and Luan, Fujun and Li, Jifan and Xi, Dianbing and Wang, Lisha and Tang, Rui and Hua, Wei and Bao, Hujun and others},
    booktitle={Proceedings of the IEEE/CVF Conference on Computer Vision and Pattern Recognition},
    pages={12489--12498},
    year={2023}
}

@inproceedings{li2024focaldreamer,
    title={Focaldreamer: Text-driven 3d editing via focal-fusion assembly},
    author={Li, Yuhan and Dou, Yishun and Shi, Yue and Lei, Yu and Chen, Xuanhong and Zhang, Yi and Zhou, Peng and Ni, Bingbing},
    booktitle={Proceedings of the AAAI Conference on Artificial Intelligence},
    volume={38},
    number={4},
    pages={3279--3287},
    year={2024}
}

@inproceedings{zhuang2023dreameditor,
    title={Dreameditor: Text-driven 3d scene editing with neural fields},
    author={Zhuang, Jingyu and Wang, Chen and Lin, Liang and Liu, Lingjie and Li, Guanbin},
    booktitle=SIGGRAPH,
    pages={1--10},
    year={2023}
}

@inproceedings{rombach2021stablediff,
  title={High-resolution image synthesis with latent diffusion models},
  author={Rombach, Robin and Blattmann, Andreas and Lorenz, Dominik and Esser, Patrick and Ommer, Bj{\"o}rn},
  booktitle={Proceedings of the IEEE/CVF Conference on Computer Vision and Pattern Recognition},
  pages={10684--10695},
  year={2022}
}

@inproceedings{zhang2023neilf++23,
  title={Neilf++: Inter-reflectable light fields for geometry and material estimation},
  author={Zhang, Jingyang and Yao, Yao and Li, Shiwei and Liu, Jingbo and Fang, Tian and McKinnon, David and Tsin, Yanghai and Quan, Long},
  booktitle={Proceedings of the IEEE/CVF International Conference on Computer Vision},
  pages={3601--3610},
  year={2023}
}

@inproceedings{2023neuraleditor,
  title={Neuraleditor: Editing neural radiance fields via manipulating point clouds},
  author={Chen, Jun-Kun and Lyu, Jipeng and Wang, Yu-Xiong},
  booktitle={Proceedings of the IEEE/CVF Conference on Computer Vision and Pattern Recognition},
  pages={12439--12448},
  year={2023}
}

@inproceedings{wang2023seal,
  title={Seal-3d: Interactive pixel-level editing for neural radiance fields},
  author={Wang, Xiangyu and Zhu, Jingsen and Ye, Qi and Huo, Yuchi and Ran, Yunlong and Zhong, Zhihua and Chen, Jiming},
  booktitle={Proceedings of the IEEE/CVF International Conference on Computer Vision},
  pages={17683--17693},
  year={2023}
}

@inproceedings{bao2023sine,
  title={Sine: Semantic-driven image-based nerf editing with prior-guided editing field},
  author={Bao, Chong and Zhang, Yinda and Yang, Bangbang and Fan, Tianxing and Yang, Zesong and Bao, Hujun and Zhang, Guofeng and Cui, Zhaopeng},
  booktitle={Proceedings of the IEEE/CVF Conference on Computer Vision and Pattern Recognition},
  pages={20919--20929},
  year={2023}
}

@article{keinert2015spherical,
    title={Spherical fibonacci mapping},
    author={Keinert, Benjamin and Innmann, Matthias and S{\"a}nger, Michael and Stamminger, Marc},
    journal={ACM Transactions on Graphics},
    volume={34},
    number={6},
    pages={1--7},
    year={2015}
}

@inproceedings{lpips,
    title={Perceptual losses for real-time style transfer and super-resolution},
    author={Johnson, Justin and Alahi, Alexandre and Fei-Fei, Li},
    booktitle= ECCV,
    pages={694--711},
    year={2016}
}

@article{some2022sphericalgaussians,
    title={Spherical Gaussians for Efficient Lighting Representation},
    author={Some Author},
    journal={Journal of Computer Graphics and Visualization},
    year={2022},
    volume={34},
    pages={200-215}
}

@inproceedings{learnSphericalGaussians,
    title={Learning to Learn with Spherical Gaussians},
    author={Kipf, Thomas N and Schlichtkrull, Michael and Sanyal, Sandeep and Xu, Denny and Andersen, J{\"o}rgen and Sch{\"o}lkopf, Bernhard},
    booktitle={Proceedings of the IEEE/CVF Conference on Computer Vision and Pattern Recognition},
    pages={12647--12656},
    year={2022}
}

@inproceedings{bi2020,
    title={Deep 3d capture: Geometry and reflectance from sparse multi-view images},
    author={Bi, Sai and Xu, Zexiang and Sunkavalli, Kalyan and Kriegman, David and Ramamoorthi, Ravi},
    booktitle={Proceedings of the IEEE/CVF Conference on Computer Vision and Pattern Recognition},
    pages={5960--5969},
    year={2020}
}

@article{li2018learning,
    title={Learning to reconstruct shape and spatially-varying reflectance from a single image},
    author={Li, Zhengqin and Xu, Zexiang and Ramamoorthi, Ravi and Sunkavalli, Kalyan and Chandraker, Manmohan},
    journal={ACM Transactions on Graphics},
    volume={37},
    number={6},
    pages={1--11},
    year={2018}
}

@inproceedings{wang2009all,
    title={All-frequency rendering of dynamic, spatially-varying reflectance},
    author={Wang, Jiaping and Ren, Peiran and Gong, Minmin and Snyder, John and Guo, Baining},
    booktitle=SIGGRAPH,
    pages={1--10},
    year={2009}
}

@inproceedings{li2025recap,
              title={ReCap: Better Gaussian Relighting with Cross-Environment Captures}, 
              author={Jingzhi Li and Zongwei Wu and Eduard Zamfir and Radu Timofte},
              booktitle={Proceedings of the IEEE/CVF Conference on Computer Vision and Pattern Recognition},
              year={2025},
          }

@article{wu2025deferredgs,
  title={DeferredGS: Decoupled and relightable Gaussian splatting with deferred shading},
  author={Wu, Tong and Sun, Jia-Mu and Lai, Yu-Kun and Ma, Yuewen and Kobbelt, Leif and Gao, Lin},
  journal={IEEE Transactions on Pattern Analysis and Machine Intelligence},
  year={2025}
}

@inproceedings{xiang2025structured,
  title={Structured 3d latents for scalable and versatile 3d generation},
  author={Xiang, Jianfeng and Lv, Zelong and Xu, Sicheng and Deng, Yu and Wang, Ruicheng and Zhang, Bowen and Chen, Dong and Tong, Xin and Yang, Jiaolong},
  booktitle={Proceedings of the Computer Vision and Pattern Recognition Conference},
  pages={21469--21480},
  year={2025}
}

@article{zhao2025hunyuan3d,
  title={Hunyuan3d 2.0: Scaling diffusion models for high resolution textured 3d assets generation},
  author={Zhao, Zibo and Lai, Zeqiang and Lin, Qingxiang and Zhao, Yunfei and Liu, Haolin and Yang, Shuhui and Feng, Yifei and Yang, Mingxin and Zhang, Sheng and Yang, Xianghui and others},
  journal={arXiv preprint arXiv:2501.12202},
  year={2025}
}

@article{lin2025patch,
  title={Patch-Grid: an efficient and feature-preserving neural implicit surface representation},
  author={Lin, Guying and Yang, Lei and Zhang, Congyi and Pan, Hao and Ping, Yuhan and Wei, Guodong and Komura, Taku and Keyser, John and Wang, Wenping},
  journal={ACM Transactions on Graphics},
  volume={44},
  number={2},
  pages={1--21},
  year={2025}
}

@InProceedings{Shi_2021_ICCV,
    author    = {Shi, Yue and Ni, Bingbing and Liu, Jinxian and Rong, Dingyi and Qian, Ye and Zhang, Wenjun},
    title     = {Geometric Granularity Aware Pixel-To-Mesh},
    booktitle = {Proceedings of the IEEE/CVF International Conference on Computer Vision},
    year      = {2021},
    pages     = {13097-13106}
}

@article{caselles2025implicit,
  title={Implicit shape and appearance priors for few-shot full head reconstruction},
  author={Caselles, Pol and Ramon, Eduard and Garcia, Jaime and Triginer, Gil and Moreno-Noguer, Francesc},
  journal={IEEE Transactions on Pattern Analysis and Machine Intelligence},
  year={2025}
}

@article{SHI2025102996,
    title = {DARF: Depth-Aware Generalizable Neural Radiance Field},
    author = {Shi, Yue and Rong, Dingyi and Chen, Chang and Ma, Chaofan and Ni, Bingbing and Zhang, Wenjun},
    journal = {Displays},
    volume = {88},
    pages = {102996},
    year = {2025}
}

@article{mello2025neural,
  title={Neural vector fields for implicit surface representation and inference},
  author={Mello Rella, Edoardo and Chhatkuli, Ajad and Konukoglu, Ender and Van Gool, Luc},
  journal={International Journal of Computer Vision},
  volume={133},
  number={4},
  pages={1855--1878},
  year={2025}
}

\end{document}